%% file: main.tex
\documentclass[10pt,twocolumn,letterpaper]{article}

\usepackage{cvpr}              

\definecolor{cvprblue}{rgb}{0.21,0.49,0.74}
\usepackage[breaklinks,colorlinks,allcolors=cvprblue]{hyperref}

\usepackage{epsfig}
\usepackage{amsmath}
\usepackage{amssymb}
\usepackage{multirow}
\usepackage{graphicx}
\usepackage{caption}
\usepackage{subcaption}
\usepackage{xcolor}
\usepackage{makecell}
\usepackage{colortbl}

\def\paperID{*****} 
\def\confName{CVPR}
\def\confYear{2026}
 
\newcommand{\sslash}{\mathbin{\mkern-3mu/\mkern-6mu/\mkern-3mu}}
\newcommand{\ourapproach}{\textsc{PICO} }

\definecolor{darkred}{HTML}{a50a0a}
\definecolor{darkblue}{HTML}{305BAB}
\definecolor{scale_decoder}{HTML}{f37910}

\usepackage{macros}

\usepackage[utf8]{inputenc} 
\usepackage[T1]{fontenc}    
\usepackage{url}            
\usepackage{booktabs}       
\usepackage{amsfonts}       
\usepackage{nicefrac}       
\usepackage{microtype}      
\usepackage{xcolor}         
\usepackage{tikz}
\usetikzlibrary{arrows.meta, positioning}
\tikzset{
  box/.style={rectangle, rounded corners, draw=orange!80, thick, fill=orange!10,
              minimum width=2.2cm, minimum height=1cm, text centered},
  arrow/.style={-{Latex[length=3mm,width=2mm]}, thick, orange!80},
}

\begin{document}

\title
{\large
\textbf{What Matters in Practical Learned Image Compression}
}

\author{
Kedar Tatwawadi\and 
Parisa Rahimzadeh\and
Zhanghao Sun\and
Zhiqi Chen\and
Ziyun Yang\and
Sanjay Nair\and
Divija Hasteer\and
Oren Rippel\and\vspace{-0.1in}
\\
\makebox[\textwidth][c]{\textit{Apple}}\\
{\tt\small oren.rippel@apple.com\vspace{-0.15in}}
}
\maketitle

\begin{abstract}
One of the major differentiators unlocked by learned codecs relative to their hard-coded traditional counterparts is their ability to be optimized directly to appeal to the human visual system. Despite this potential, a perceptual yet practical image codec is yet to be proposed.

In this work, we aim to close this gap. We conduct a comprehensive study of the key modeling choices that govern the design of a practical learned image codec, jointly optimized for perceptual quality and runtime~---~including within the ablations several novel techniques. We then perform performance-aware neural architecture search over millions of backbone configurations to identify models that achieve the target on-device runtime while maximizing compression performance as captured by perceptual metrics. 

We combine the various optimizations to construct a new codec that achieves a significantly improved tradeoff between speed and perceptual quality. Based on rigorous subjective user studies, it provides 2.3-3× bitrate savings against AV1, AV2, VVC, ECM and JPEG-AI, and 20-40\% bitrate savings against the best learned codec alternatives. At the same time, on an iPhone 17 Pro Max, it encodes 12MP images as fast as 230ms, and decodes them in 150ms~---~faster than most top ML-based codecs run on a V100 GPU. 

\end{abstract}

\vspace{-0.2in}
\section{Introduction}
\vspace{-0.05in}
Since their emergence \cite{balle2016opt,rippel17,balle2018variational}, learned image codecs have shown meaningful compression gains over traditional codecs. In recent years, the field has made significant progress in addressing several challenges that had once hindered practical deployment~---~improving computational efficiency, achieving fine-grained rate control with minimal overhead, and ensuring reliable cross-platform coding which is not inherent to hyperprior-based codecs \cite[\eg][]{balle2018integer,rippel2021elf,tian2023towards,pang2024towards,jia2025towards,jpegaiRepo}. A major milestone in this evolution is the standardization of JPEG-AI \cite{jpegaiRepo}, which not only highlights the technical maturity of learned codecs but also their growing industrial traction, signaling a clear transition beyond academic research.

\begin{figure}[b]
    \vspace{-0.1in}
    \centering
    \includegraphics[width=\linewidth]{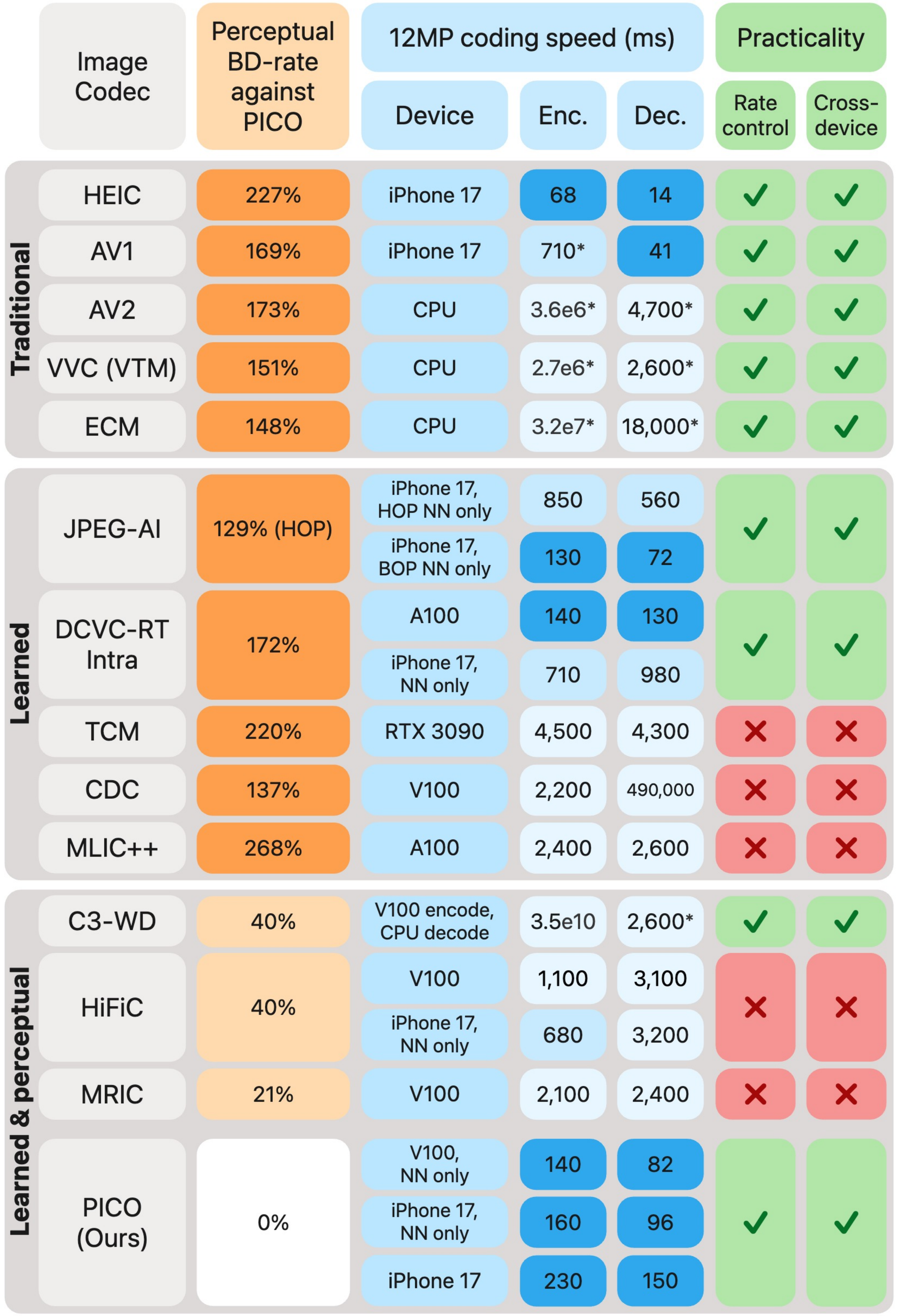}
    \vspace{-0.2in}
    \caption{Comparisons of state-of-the-art traditional and learned codecs across different considerations of practicality. The reported perceptual BD-rates are based on human ratings from a large-scale subjective study (Sec.~\ref{sec:results}). For speed comparisons on iPhone 17 Pro Max, we use the exact architecture implementations found in the repositories of the baselines, and apply the same compiler optimizations as for PICO. Benchmarks marked with $^*$ indicate that the runtime is expected to be faster once accelerated in hardware.}
    \label{fig:spotlight_figure}
\end{figure}

\begin{figure*}[t!]
    \centering
    \includegraphics[width=1\linewidth]{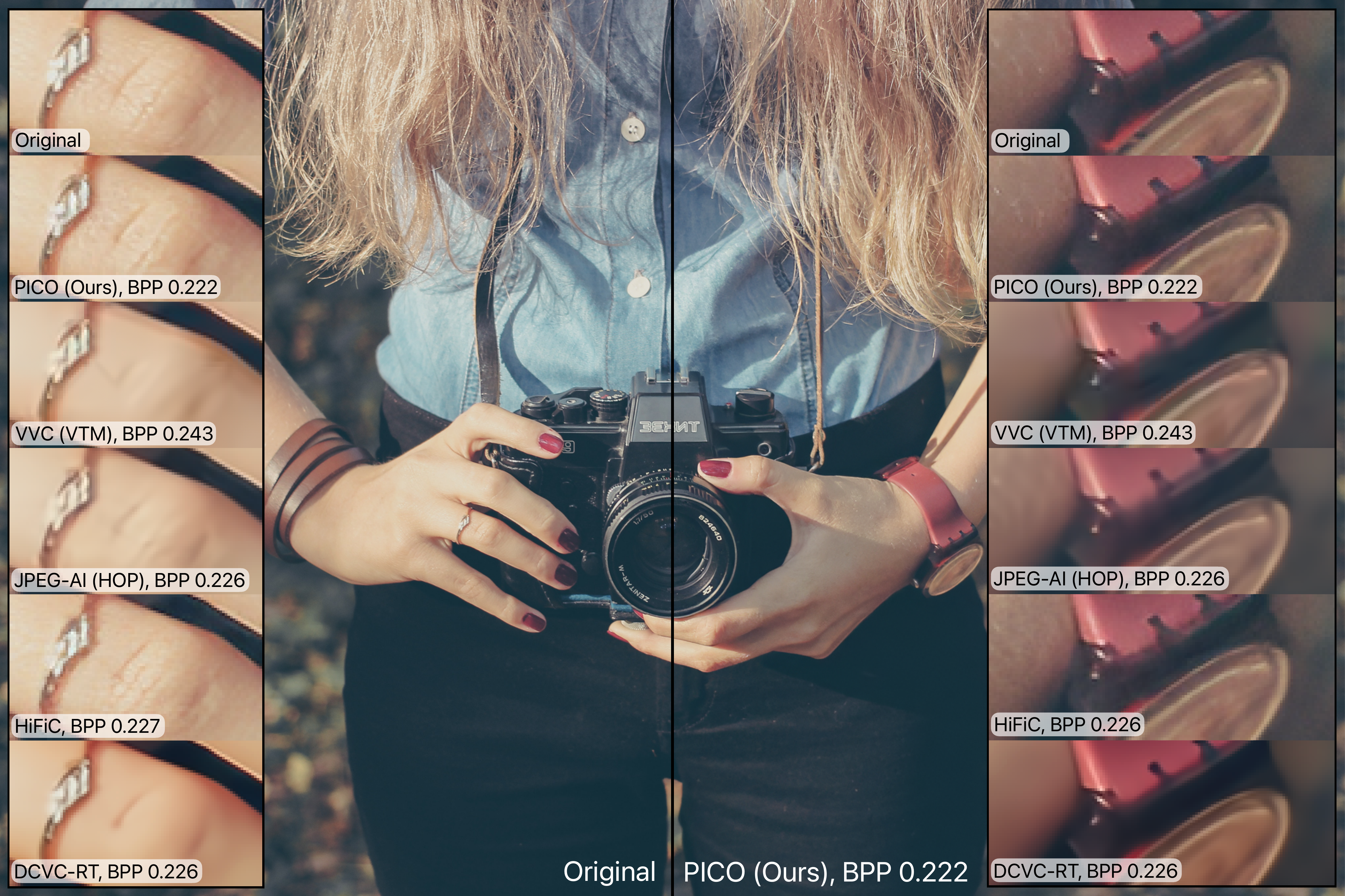}
    \caption{Qualitative comparisons of reconstruction quality for equal filesize/BPP (bits-per-pixel). \ourapproach features significant improvements to fine-grained detail preservation, and even at low bitrates remains indistinguishable from the original.}
    \label{fig:qualitative_comparison}
    \vspace{-0.15in}
\end{figure*}

Despite these remarkable advancements in building learned image codecs, a major opportunity remains largely untapped. The key advantage of learned codecs over traditional hand-engineered approaches lies in their ability to be directly optimized for the task at hand~---~which is often to appeal to the human visual system. Several studies have explored this direction, establishing the foundations for applying modern generative techniques to image compression \cite{mentzer2020high,yang2023lossy,balle2025good,he2022poelicperceptionorientedefficientlearned}. Although these works have demonstrated the exciting potential for perceptual optimization, their runtimes are an order of magnitude away from practical deployment. Moreover, most of them lack features necessary for any practical codec, such as cross-platform support or rate control. 

In this work, we aim to close this gap. Our key contributions are as follows: 
\begin{itemize} 
\item We present the first work to comprehensively ablate across a broad spectrum of modeling decisions, and millions of model configurations, to explicitly optimize the trade-off between perceptual quality and runtime. The ablations include several novel architectures and algorithmic techniques, aimed at maximizing the codec's expressivity~---~crucial for its generative capability~---~while explicitly avoiding incurring computational overhead.
\item We introduce carefully-designed training and loss recipes that enable stable optimization of lightweight codecs towards high perceptual quality. We further propose specialized losses to surgically mitigate text and tiling artifacts.
\item Building on these systematic ablations, we introduce \textbf{\ourapproach (\underline{P}erceptual \underline{I}mage \underline{Co}dec)}, a new image codec that integrates all essential components for practical deployment. Through extensive subjective user studies, PICO achieves 2.3–3× bitrate savings over AV1, AV2, VVC, ECM, and JPEG-AI, and 20–40\% savings compared to the strongest learned codec baselines (Fig.~\ref{fig:spotlight_figure},~\ref{fig:qualitative_comparison},~\ref{fig:perceptual_rd_curves}). On an iPhone 17 Pro Max, PICO encodes 12MP images in as little as 230ms and decodes them in 150ms~---~faster than most state-of-the-art learned codecs run on a V100 GPU. 
\end{itemize}

\begin{figure*}
    \centering
    \includegraphics[width=1\linewidth]{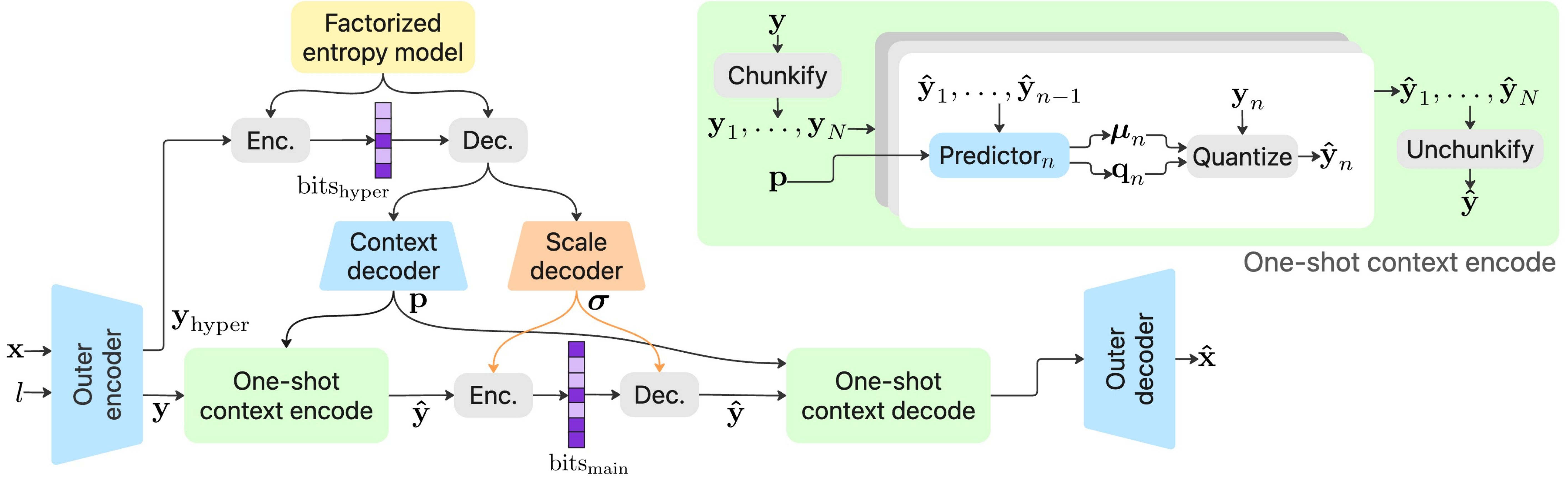}
    \vspace{-0.2in}
    \caption{The overall model architecture. Individual components described in Sections \ref{sec:codec_framework} and \ref{sec:large_model_changes}. The \textbf{\textcolor{scale_decoder}{scale decoder}} computation is bit-exact to guarantee entropy decodability.}
    \label{fig:high_level_architecture}
    \vspace{-0.15in}
\end{figure*}

\vspace{-0.05in}
\section{Related work}
\vspace{-0.05in}
\noindent\textbf{Traditional image codecs} such as BPG~\cite{bpg_source_code}, VVC~\cite{vvc_source_code}, AV1~\cite{av1_source_code} and next-generation ECM~\cite{ecm_source_code} and AV2~\cite{av2_source_code} are based on hand-crafted pipelines that exploit redundancy by combining transformations with entropy coding. While these codecs have been extensively optimized, their design is fundamentally constrained by heuristically-designed components, leading to several limitations. For example, although they can be slightly tuned towards given metrics, their structure makes it inherently challenging to explicitly optimize them for perceptual quality. They also typically require dedicated hardware, leading to long adoption and update cycles.

\vspace{0.05in}
\noindent\textbf{Learned image codecs} aim to resolve these issues via end-to-end modeling using neural networks~\cite{balle2016opt,rippel17,balle2018variational}, allowing them to be explicitly optimized to achieve optimal tradeoffs between bitrate, and given differentiable metrics. This unlocked the ability to train codecs directly for perceptual quality~\cite{agustsson2018generative, mentzer2020high,agustsson2023multi,jia2024generative,balle2025good}. Recent research~\cite{yang2023lossy, relic2024bridging} proposes to employ latent diffusion for image compression.

\vspace{-0.15in}
\paragraph{Practical learned image compression} Despite their promise, learned image codecs have faced several major challenges. First, achieving high perceptual quality requires the model to align with the human visual system. Prior art introduced perceptual training objectives~\cite{mentzer2020high,balle2025good, cool-chic}, yet still produce noticeable artifacts. Second, practical on-device deployment scenarios demand fast encoding and decoding. Many learned codecs (including all perceptual codecs mentioned) rely on heavyweight neural architectures~\cite{yang2023lossy,jia2024generative}, autoregressive entropy models~\cite{minnen2018joint, he2021checkerboard, he2022elic, li2024neural}, or test-time optimization~\cite{kim2024c3,balle2025good} to enhance compression efficiency~---~at the cost of computational overhead. Recent research proposes more efficient neural architectures~\cite{jia2025towards, jpegaiRepo} but focuses on metrics such as PSNR or SSIM, which poorly reflect perceptual quality. Third, encoding/decoding across devices with differing hardware/software configurations needs to be supported. Proposed enablements include integer-only coding to avoid inherent non-determinism in floating point operations~\cite{balle2018integer}, vector quantization to avoid decoding failures~\cite{lu2019learning}, and additional signaling to safeguard against errors \cite{pang2024towards}.

\vspace{-0.05in}
\section{Codec framework}
\label{sec:codec_framework}
\vspace{-0.05in}

Before diving into the details of the codec design search space, we first describe the framework at a high level. 

\begin{figure*}[t]
    \centering
    \includegraphics[width=0.96\textwidth]{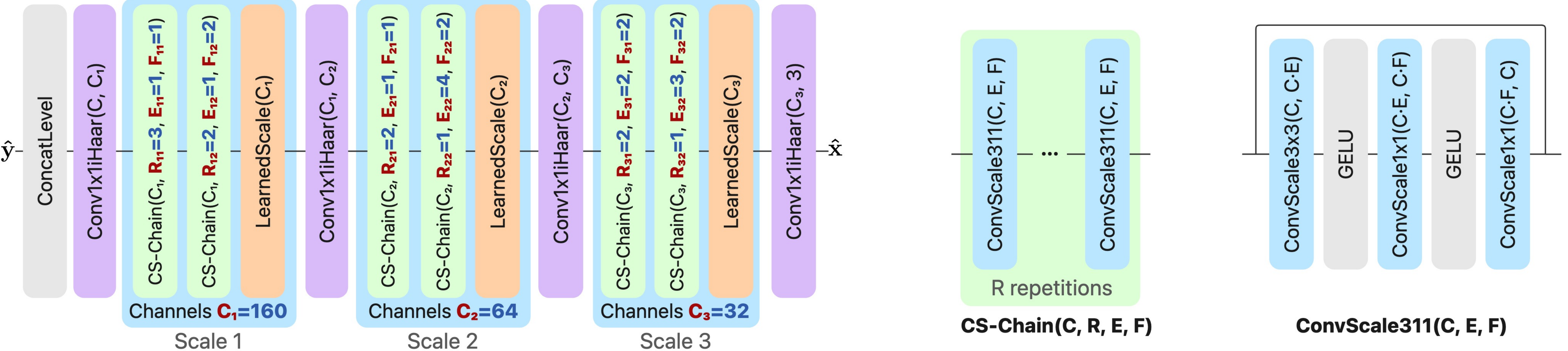}
    \vspace{-0.05in}
    \caption{Detailed architecture of the outer decoder (see Appendix~\ref{app:architecture} for specifications of other model components). \textbf{Left}: We searched over millions of configurations from this model family, as defined by the \textbf{\textcolor{darkred}{hyperparameters in red}} with \textbf{\textcolor{darkblue}{optimal values in blue}}, to achieve target iPhone runtimes while maximizing perceptual compression efficiency (see Sec.~\ref{sec:searching_design_space}). \textbf{Right}: The architecture of the $\textrm{ConvScale311}(C, E, F)$ module, with $C$ channels, and expansion factors $E, F$. The base ConvScale layer is a reparametrization of a convolution with additional learned scales, and is described in Sec.~\ref{sec:large_model_changes}. \textbf{Middle}: the $\textrm{CS-Chain}(C, R, E, F)$ module simply repeats this block $R$ times.}
    \label{fig:detailed_decoder}
    \vspace{-0.15in}
\end{figure*}

\subsection{High-level codec framework}
\vspace{-0.05in}
The ubiquitous hyperprior architecture described in \cite{minnen2018joint} includes four sub-networks: encoder, decoder, hyper-encoder, and hyper-decoder. The encoder and decoder networks are responsible for converting the input image $\rmbx$ to a latent tensor $\rmbhy$ and back to a reconstruction $\rmbhx$, while the hyper-encoder and hyper-decoder are used to provide parameters for entropy coding of the latent tensor $\rmbhy$. Specifically, the hyper-decoder outputs parameters location $\bmu$ and scale $\bsigma$, which are used by the entropy coder to map to a discrete distribution for lossless coding of the latent $\rmbhy$. 

At a high level, our model framework is similar to the hyperprior architecture, albeit with a few key differences. First, we split the hyper-decoder network into two sub-networks: a \emph{scale decoder} and a \emph{context decoder} (Fig.~\ref{fig:high_level_architecture}). The scale decoder outputs the scale parameter $\bsigma$ used for entropy coding of the latent $\rmbhy$ and hence must produce the \textit{exact same} output during the encoding and decoding processes given the extreme sensitivity of entropy decoding to parameter mismatch. Separating the scale decoder into a standalone model is crucial in facilitating guaranteed cross-device robustness, as well as unlocking additional speed gains via pipelining~(Sec.~\ref{sec:extensions_practicality}). The context decoder can be thought of as a generalization of the location~$\bmu$ output from the hyper-decoder model (see Sec.~\ref{sec:searching_design_space}). Another key difference is that the hyper-encoder network is absorbed into the encoder network (Fig. \ref{fig:high_level_architecture}). This simplification allows for the encoder to be compiled and executed as a single network. 

\vspace{-0.05in}
\subsection{Extensions for practical deployment}
\label{sec:extensions_practicality}
\vspace{-0.05in}
We further extend this model in several ways to allow for practical deployment:

\vspace{-0.15in}
\paragraph{Guaranteed cross-platform robustness} As was observed in many prior works \cite[\eg][]{balle2018integer,tian2023towards,pang2024towards,jia2025towards}, the parameters provided to the entropy coder must be \emph{bit-exact}: the slightest discrepancy in computation between the entropy encoder and decoder will result in decoding failure. To guarantee success, we build the scale decoder to provide deterministic output across devices. We first quantize the model to UINT8 so that all the weights and activations within the network are integers. This step is necessary but in fact not sufficient, as there remain some floating point (FP) operations through the quantization scaling factors. Though these FP operations cannot be reordered by the compiler~---~a primary culprit for nondeterministic output~---~we cannot be sure how different hardware architectures may handle the FP arithmetic (\ie precision and rounding modes). Thus, to achieve cross-platform determinism, we opt to run the scale decoder on CPU for compliance with the IEEE FP standard.

\vspace{-0.15in}
\paragraph{Quality level control} \label{par:quality_control} We use a single model to represent the entire bitrate range, at negligible costs to both computation and model size. To do so, we condition the encoder and decoder networks, as well as loss definitions, on a scalar quality level $l$ signaled in the bitstream. We follow the level embedding recipe described in Appendix~E of \cite{rippel2021elf} as our starting point, to which we apply several enhancements. The details can be found in Appendix~\ref{app:quality_level_control}.

\begin{figure*}
    \centering
    \includegraphics[width=1\linewidth]{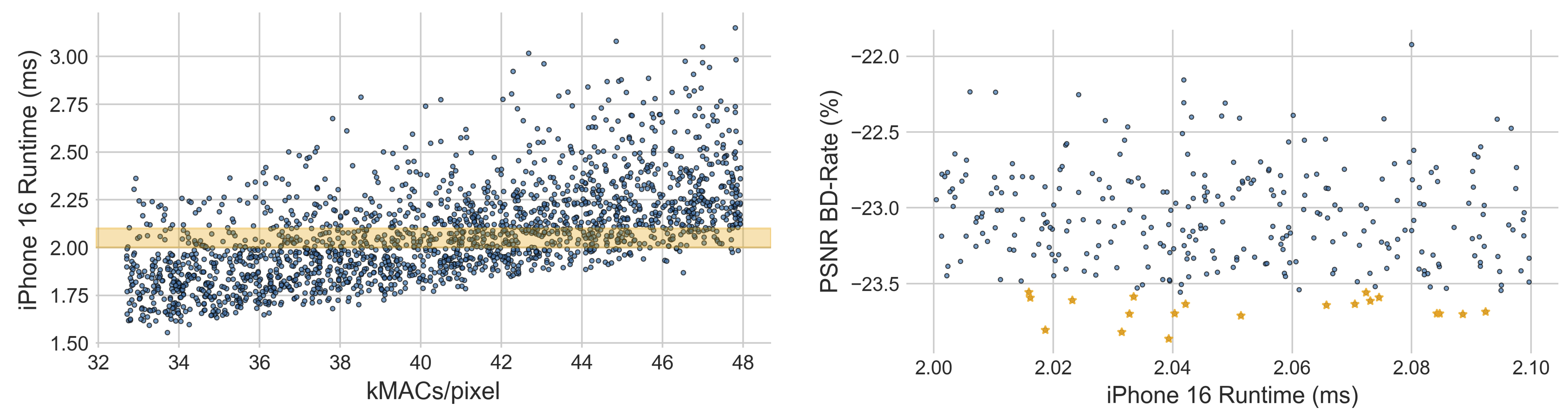}
    \vspace{-0.3in}
    \caption{We perform neural architecture search for the outer decoder, progressively filtering the search space down from 1.4M model candidates to 20 models which are trained to completion (Sec.~\ref{sec:neural_architecture_search}). Note that the runtime reflects the time taken to decode a single $512\times512$ tile. \textbf{Left}: We benchmark the runtimes of 10,000 decoder candidates on-device, and show kMACs/pixel vs. iPhone 16 Pro runtimes (visualized for a sample of 2k models). These are further filtered by runtime, range highlighted in yellow, to choose a subset of 1,000 models to perform partial-training based filtering. \textbf{Right}: On-device runtime vs. PSNR BD-Rate for the 1,000 models trained (small subset visualized). Highlighted are the final shortlisted 20 models chosen to train to completion using the full perceptual recipe.}
    \label{fig:hp_sweep}
    \vspace{-0.15in}
\end{figure*}

\vspace{-0.15in}
\paragraph{Tile processing and pipelining} 
\label{sec:tiling}
We introduce spatial tiling to improve computational efficiency. This enables pipelined execution, where the entropy coding and scale decoding of one tile run on the CPU while the neural components of another tile run concurrently on the accelerator. Each image is partitioned into non-overlapping tiles of size $504\times504$. During encoding, each tile is padded to $512\times512$ with a 4-pixel contextual margin sourced from neighboring tiles. Including neighboring context helps maintain feature continuity across the tile boundary,  partially mitigating tiling artifacts. Residual inconsistencies are further reduced by incorporating training losses which emphasize consistency across independent tile reconstructions (see Section~\ref{sec:training_loss_enhancements}).

\vspace{-0.05in}
\subsection{Loss \& training procedure}
\label{sec:loss_training}
\vspace{-0.05in}
\paragraph{Loss} Our combined rate-distortion loss function used for training is as described by Eq.~\ref{eqn:total_loss}. Similar to other perceptual-oriented learned codecs \cite[\eg][]{mentzer2020high, agustsson2018generative, agustsson2023multi}, we use a combination of pixel-matching losses, perceptual losses, GAN-based losses, and losses to surgically mitigate specific artifacts. We ablate on different choices in detail in Section~\ref{sec:training_loss_enhancements}.

\vspace{-0.15in}
\paragraph{Training procedure} 

We adopt the following training procedure for all experiments. The codec is trained on an internal dataset comprising approximately 90k generic images, analogous to ImageNet, supplemented with 2.3k images of text content, and another 28k high resolution open-sourced dataset from Div2K \cite{Agustsson_2017_CVPR_Workshops}, CLIC \cite{clic-challenge}, and Flickr2K \cite{flickr2k}. We use the Adam optimizer \cite{kingma2014adam}. The training is split into two phases: to start, the codec is trained solely on MSE; afterwards, the various perceptual losses introduced (see Sec.~\ref{sec:training_loss_enhancements} and Appendix~\ref{app:pq_training} for further detail).

\vspace{-0.05in}
\section{Studying the codec design space}
\label{sec:searching_design_space}
\vspace{-0.05in}
We comprehensively explore the codec design space, specifically focusing on directions that would not increase computational complexity. We explore large architectural changes in Section~\ref{sec:large_model_changes}; perceptual optimizations in \ref{sec:training_loss_enhancements}; and comprehensively search over how to best configure the backbone hyperparameters in \ref{sec:neural_architecture_search}. For all these experiments, we keep the training procedure (end of Sec.~\ref{sec:codec_framework}) constant.

\vspace{-0.05in}
\subsection{Model Architecture enhancements}
\label{sec:large_model_changes}
\vspace{-0.05in}
We present in detail modeling enhancements that are geared towards obtaining improved expressivity and capacity without impact on speed. Each enhancement is separately validated in the ablation studies (Sec.~\ref{sec:findings} and Tab.~\ref{tab:network_architecture_ablation}).

\vspace{-0.18in}
\paragraph{Backbone and learned scales} Our starting point for the backbone of the encoder/decoder models is an inverted residual \cite{sandler2018mobilenetv2} with several modifications, which we call \mbox{\textbf{ConvScale311}} (Fig.~\ref{fig:detailed_decoder}). As validated by ablation studies, it provides a strong tradeoff between computational efficiency and expressivity. The architecture features different types of learned elementwise scales, which we find significantly improve the stability and performance of the model, at a negligible computational overhead:
\begin{enumerate}
    \item Consider a convolution with $C$ input channels, $K$ output channels, $G$ groups, and kernel size $Y\times X$ with weight $\rmbW$ and bias $\rmbb$ of sizes $[K, C \sslash G, Y, X]$ and $[K]$. We define a new variant of the convolution layer we call \textbf{ConvScale}, which we supplement with two additional learned parameters: an input scale $\rmbs_{\textrm{in}}$ and output scale $\rmbs_{\textrm{out}}$ with shapes $[1, C\sslash G, 1, 1]$ and $[K, 1, 1, 1]$. We parameterize the weight and bias to explicitly learn the scales as $\rmbW'=\rmbs_{\textrm{in}}\rmbs_{\textrm{out}}\rmbW$ and $\rmbb' = \textrm{squeeze}(\rmbs_{\textrm{out}})\rmbb$. During inference we reparameterize $\rmbW'$ and $\rmbb'$ by collapsing the scales into them, leading to identical computational costs as for a normal convolution. We use ConvScale in place of all convolutions in the model.
    \item We further introduce learned elementwise scaling factors that modulate activations near the end of each processing block corresponding to each spatial resolution (Fig.~\ref{fig:detailed_decoder}).
\end{enumerate}

\begin{figure*}[t!]
    \centering
    \includegraphics[width=1\linewidth]{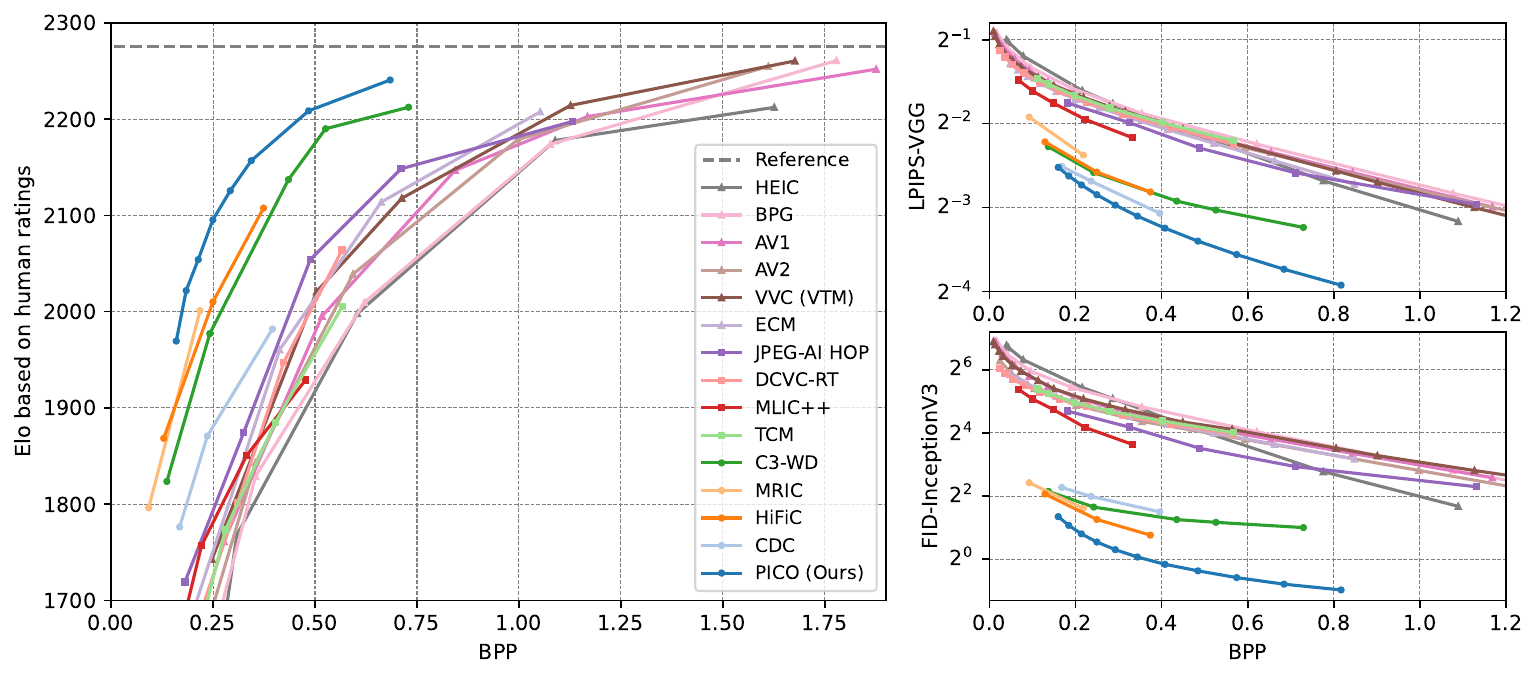}
    \vspace{-0.35in}
    \caption{Rate-distortion curves of top traditional and learned codecs, based on Elo scores (higher is better) from a large-scale subjective study, and perceptual objective metrics (lower is better) on the CLIC 2020 test dataset. Traditional codecs are indicated with $\blacktriangle$ markers, learned codecs with $\blacksquare$, and perceptual+learned codecs with {\large $\bullet$}. Evaluations on additional metrics and datasets can be found in Appendix~\ref{app:additional_evaluations}.}
    \label{fig:perceptual_rd_curves}
    \vspace{-0.15in}
\end{figure*}

\vspace{-0.15in}
\paragraph{Learned quantization width} It is a common methodology in learned compression for the hyperprior decoder to predict an elementwise location parameter $\bmu$ to shift the distribution used to code $\rmby$ \cite{minnen2018joint}. In our work, this is accomplished by the context decoder (Sec.~\ref{sec:codec_framework}); in addition, we find that it is helpful for the context decoder to also produce an input-specific elementwise learned quantization width $\textbf{\textrm{q}}>0$ to adaptively modulate the width of the quantization bins. In practice, the context decoder (Fig.~\ref{fig:high_level_architecture}) produces a prior $\rmbp$ which is then mapped to $\bmu, \rmbq$ by the context model (see below). We then quantize the main latent by rounding to the nearest integer $\rmbhy=\lfloor\frac{\rmby - \bmu}{\rmbq}\rceil$, which we then entropy-encode. After entropy-decoding, we invert the operations as $\rmbq\rmbhy + \bmu$.

\vspace{-0.18in}
\paragraph{One-shot context model} While learned codecs benefit significantly from autoregressive (AR) coding \cite[\eg][]{minnen2018joint,he2021checkerboard,he2022elic,li2024neural}, it results in slowdowns due to repeated back-and-forth memory transfers between the CPU and ML accelerator as entropy coding is interlaced with prediction. We observe, however, that this shortcoming is only a product of applying AR specifically to the \emph{scale} $\bsigma$ which is required for entropy decoding. That is, if we decode the scale in a one-shot fashion, then we can freely apply iterative AR strategies to $\bmu, \rmbq$ while keeping the computation exclusively on the ML accelerator. We refer to this as a \emph{one-shot context model} (Fig. $\ref{fig:high_level_architecture})$, which enjoys the benefits of AR at a negligible speed penalty. The iterative prediction structure can be chosen analogously to true AR: for example, as channel-wise steps \cite{minnen2020channel}, checkerboard \cite{he2021checkerboard}, and so on. We note that JPEG-AI \cite{jpegaiRepo} independently developed a component in a similar spirit, albeit applied to the $\bmu$ only, and with twice the AR prediction steps.

\vspace{-0.18in}
\paragraph{Conv + Haar Resampling}
Motivated by the Cosmos tokenizer~\cite{nvidia2025cosmosworldfoundationmodel}, we employ 2D Haar wavelets for all resampling operations in the codec. Haar wavelets decompose the input into partially de-correlated channels in an invertible manner, with an analogous inverse transform. This can be interpreted as imposing an inductive bias on each learned resampling operation, promoting structured multi-scale representations and effectively increasing model capacity. 

In this work, we introduce a reparametrization trick to add Haar/iHaar wavelets into the codec at \emph{zero additional} computational cost; see Appendix~\ref{app:haar} for full details.

\subsection{Training loss enhancements}
\label{sec:training_loss_enhancements}
\vspace{-0.05in}
Keeping the model architecture constant, we notice that difference in training loss could lead to significant improvements to the model performance. Our cumulative distortion loss term used to train \ourapproach is as follows: 
\vspace{-0.1in}
\begin{eqnarray}
\label{eqn:distortion_loss}
D = \text{MSE}(\rmbx, \rmbhx) + w_1~ \text{LPIPS}(\rmbx, \rmbhx) + w_2~ \text{MS-SSIM}(\rmbx, \rmbhx) \nonumber
\\
{\small + w_3~ \textrm{TilingArtifactLoss}(\rmbx, \rmbhx) + w_4~ \textrm{TextFidelityLoss}(\rmbx, \rmbhx, \rmbm)}\nonumber \\
+ w_5~ \text{GAN}(\rmbx, \rmbhx, \rmbm)
\nonumber
\end{eqnarray}
\vspace{-0.05in}
We describe the rationale behind each loss term below.

\vspace{-0.15in}
\paragraph{Pixel-matching \& perceptual losses} In general, while the GAN significantly improves the visual realism, we notice that without appropriate pixel-matching + perceptual terms, it generates artifacts and hallucinates details. We moreover observe that a combination of pixel-matching and perceptual losses (MSE, LPIPS \cite{zhang2018unreasonable}, MS-SSIM \cite{wang2003multiscale}) allows for better regularization of the GAN, as it can no longer exploit specific weaknesses within a single loss. 

\vspace{-0.18in}
\paragraph{Text artifact mitigation} The human visual system is extremely sensitive to distortions to text, where even the smallest hallucinations would render it unreadable. To this end, we augment the perceptual training with the \mbox{\emph{TextFidelityLoss}} term. We use an off-the-shelf text detector~\cite{baek2019character} to generate a saliency mask. In the salient regions, a heavy L1 loss is then applied, while the GAN-based losses are subdued. In Section~\ref{sec:pq_ablation} we show the effectiveness of this approach.

\vspace{-0.18in}
\paragraph{Low-frequency tiling artifact mitigation} \ourapproach runs in a tiled fashion (Section~\ref{sec:tiling}), which leads to tiling artifacts in the absence of targeted mitigation. Specifically, perceptual losses and GAN in general ignore low spatial frequency components in the reconstruction, leading to color mismatch between neighboring tiles. To this end, we introduce \mbox{\emph{TilingArtifactLoss}} (TAL), a multi-resolution L1 loss which imposes fidelity supervision on multiple spatial frequencies. We show ablations of this loss term in Section~\ref{sec:pq_ablation}.

\vspace{-0.18in}
\paragraph{GAN training \& discriminator design}
Consistent with the observations of \cite{mentzer2020high, agustsson2023multi}, we find that GAN-based training significantly improves the perceptual quality. Typically, a stronger discriminator provides better supervision to the generator (the codec), resulting in improved generation quality. We use a patch-wise discriminator architecture similar to \cite{isola2017image}, but boost the discriminator capacity by increasing the number of channels and convolution layers.

However, a larger discriminator leads to training instabilities, given that the lightweight decoder has limited capacity. We employ various strategies to stabilize GAN training. First, we utilize a two-stage training recipe. The first stage uses MSE as the only distortion loss. In the second stage, the perceptual fine-tuning stage, all distortion loss terms in Eq.~\ref{eqn:distortion_loss} are added to optimize the perceptual quality. This approach improves stability by allowing the GAN-based training to start with a reasonable initialization. We also follow a warm-up schedule by gradually increasing the weight of discriminator supervision as the training proceeds. This mitigates the risk of the compression model being misled while the discriminator is in the early stage of training. 

\vspace{-0.05in}
\subsection{Neural architecture search}
\label{sec:neural_architecture_search}
\vspace{-0.05in}
On top of the high-level modeling decisions introduced in Section~\ref{sec:large_model_changes}, we further conduct neural architecture search (NAS) to optimize over the large space of backbone hyperparameter (HP) choices. We search for models that maximize compression performance, while abiding by a target on-device runtime. We describe the process we followed for the decoder NAS; we follow similar processes for the other sub-models with details found in Appendix~\ref{app:nas}.

We optimize over the decoder model family presented in Fig.~\ref{fig:detailed_decoder} with the NN runtime target of 100ms for a 12MP image on an iPhone 16 Pro. This runtime threshold was chosen as the decoding speeds acceptable for real-life use.  Na\"{i}vely taking the Cartesian product of the value sets for each HP results in $\sim$1.4M candidate models. Given the huge number of candidate models, we proceed systematically to narrow the search space in a multi-step filtering process:
\begin{enumerate}
    \item \textbf{kMACs/pixel filtering}: Given that computing operation counts is cheap, we use kMACs/pixel as a coarse form of filtering to eliminate candidates that are clearly out of bounds. Based on a preliminary analysis of typical runtimes as function of operation counts, we filter out any models with kMACs/pixel counts outside of $[32.7, 48.0]$, reducing the search space to $\sim$500k candidates.
    \item \textbf{On-device runtime filtering}: Since MACs only loosely reflect runtime (see Fig.~\ref{fig:hp_sweep}), we benchmark the actual runtimes of randomly-sampled 10k models on an iPhone 16 Pro and filter models more than 5\% away from the target runtime, resulting in $\sim$1,000 models.
    \item \textbf{Compression performance filtering}: To reduce computational cost, we partially train the selected models for the first phase only (Sec.~\ref{sec:loss_training}), and for 30\% of the epochs. The results can be found in Fig.~\ref{fig:hp_sweep}. We choose the top 20 models based on PSNR BD-rate.
    \item \textbf{Full training of the final candidates}: Finally, we train the 20 models fully, and pick the top model based on performance on perceptual metrics and visual evaluation.
\end{enumerate}

In Appendix \ref{app:nas}, we discuss the discovered architecture and provide intuition on why it provides a good tradeoff between capacity and speed. The encoder/decoder respectively have 15.2M/9.6M parameters, are 30.4MB/19.4MB on disk, and have peak memory use of 38.8MB/25.4MB on device.

\vspace{-0.05in}
\section{Results}
\label{sec:results}
\vspace{-0.05in}
We consolidate insights from our exploration of the codec design space to develop PICO~---~a practical learned image codec optimized for alignment with human perception. In this section, we evaluate PICO’s performance in depth.

\vspace{-0.05in}
\subsection{Evaluation procedure}
\label{sec:experimental_procedure}
\paragraph{Datasets} We evaluate all the codecs on the commonly-used CLIC 2020 Test dataset~\cite{clic-challenge}, consisting of 428 images of varying resolutions. In Appendix \ref{app:additional_evaluations}, we share subjective and objective results on the Kodak and DIV2K \cite{Agustsson_2017_CVPR_Workshops} datasets.

\vspace{-0.15in}
\paragraph{Baselines} We comprehensively compare to state-of-the-art codecs; their specific configurations can be found in Appendix~\ref{app:baselines}. From the traditional codecs, we compare to HEIC, and the reference implementations of BPG \cite{bpg_source_code}, AV1 \cite{av1_source_code}, VVC (VTM) \cite{vvc_source_code} and of next-generation codecs AV2 \cite{av2_source_code} and ECM \cite{ecm_source_code}. In terms of learned codecs, we compare to HiFiC \cite{mentzer2020high}, JPEG-AI \cite{jpegaiRepo}, MLIC++ \cite{jiang2023mlicpp}, CDC \cite{yang2023lossy}, TCM \cite{liu2023learned}, MRIC \cite{agustsson2023multi}, C3-WD \cite{balle2025good,cool-chic}, and DCVC-RT \cite{jia2025towards}. For JPEG-AI, we evaluate the quality of the stronger-but-slower High Operation Point (HOP), and for completeness share speed benchmarks of also the Base Operation Point (BOP).

\vspace{-0.15in}
\paragraph{Metrics} In this work, we focus exclusively on perceptual quality, and as such report on popular perceptually-aligned metrics: CMMD \cite{jayasumana2024rethinking}, FID \cite{heusel2017gans} and LPIPS \cite{zhang2018unreasonable}. We report PSNR results in Appendix~\ref{app:additional_evaluations}, and observe that it poorly reflects perceptual quality~---~a well-known shortcoming.

\begin{table}[b]
\vspace{-0.1in}
\footnotesize
\centering
\begin{tabular}{llc}
\noalign{\hrule height 1pt}
Property ablated & Option & \makecell{CMMD-CLIP \\ BD-Rate} \\
\noalign{\hrule height 1pt}
\multirow{4}{*}{\makecell[l]{One-shot\\ autoregressivity}} 
& None & \cellcolor{red!45}10.28\% \\
& Channel-wise (4 groups) & \cellcolor{red!30}3.10\% \\
& Checkerboard & \cellcolor{red!45}14.67\% \\
\cline{2-3}
& 2x2 grid & 0\% \\
\noalign{\hrule height 1pt}
\multirow{2}{*}{\makecell[l]{Learned\\ quantization width}} 
& No$^*$ & \cellcolor{red!45}8.16\% \\
\cline{2-3}
& Yes & 0\% \\
\noalign{\hrule height 1pt}
\multirow{4}{*}{Learned Scale}
& None$^{**}$ & \cellcolor{red!45}9.58\% \\
& ConvScale only$^*$ & \cellcolor{red!30}3.76\% \\
& Per spatial scale only$^*$ & \cellcolor{red!15}1.21\% \\
\cline{2-3}
& ConvScale + per spatial scale  & 0\% \\
\noalign{\hrule height 1pt}
\multirow{3}{*}{\makecell[l]{Resampling}} 
& Pixel reshuffling & \cellcolor{red!45}19.51\% \\
& Stride-2 Conv \& Deconv$^*$ & \cellcolor{red!45}8.90\% \\
\cline{2-3}
& Haar-based resampling & 0\% \\
\noalign{\hrule height 1pt}
\multirow{2}{*}{\makecell[l]{All above properties}} 
& Disabled & \cellcolor{red!60}31.69\% \\
\cline{2-3}
& Enabled & 0\% \\
\noalign{\hrule height 1pt}
\end{tabular}
\caption{Architectural ablations, as evaluated on the CLIC 2020 testset. For each property, the BD-rate was computed with the anchor being the final chosen setting, in the last row. Every~$^*$ indicates halving of the learning rate to stabilize training.\vspace{-0.1in}}
\label{tab:network_architecture_ablation}
\end{table}
\begin{table}[b]
\footnotesize
\centering
\begin{tabular}{llcl}
\noalign{\hrule height 1pt}
Evaluation metric & Property ablated & Option & Value \\
\noalign{\hrule height 1pt}
\multirow{2}{*}{\makecell[l]{L1 in text regions}} & 
\multirow{2}{*}{\makecell[l]{Text fidelity loss }}& Off & \cellcolor{red!30} 0.0093 \\
\cline{3-4}
& & On & \ 0.0046 \\
\noalign{\hrule height 1pt}
\multirow{2}{*}{\makecell[l]{Low-frequency error \\ across tile boundaries}} & 
\multirow{2}{*}{\makecell[l]{Tiling artifact loss}} & 
Off & \cellcolor{red!30} 0.0020 \\
\cline{3-4}
& & On & \ 0.00097 \\
\noalign{\hrule height 1pt}
\end{tabular}
\vspace{-0.1in}
\caption{Artifact-specific loss ablations, as evaluated by specific metrics constructed to quantify the artifacts as described in Sec.~\ref{sec:training_loss_enhancements}. Visual examples can be found in Fig.~\ref{fig:ablation_pq}.}
\label{tab:loss_ablation}
\vspace{-0.15in}
\end{table}

\begin{figure*}[t!]
\begin{center}
\includegraphics[width=0.95\linewidth]{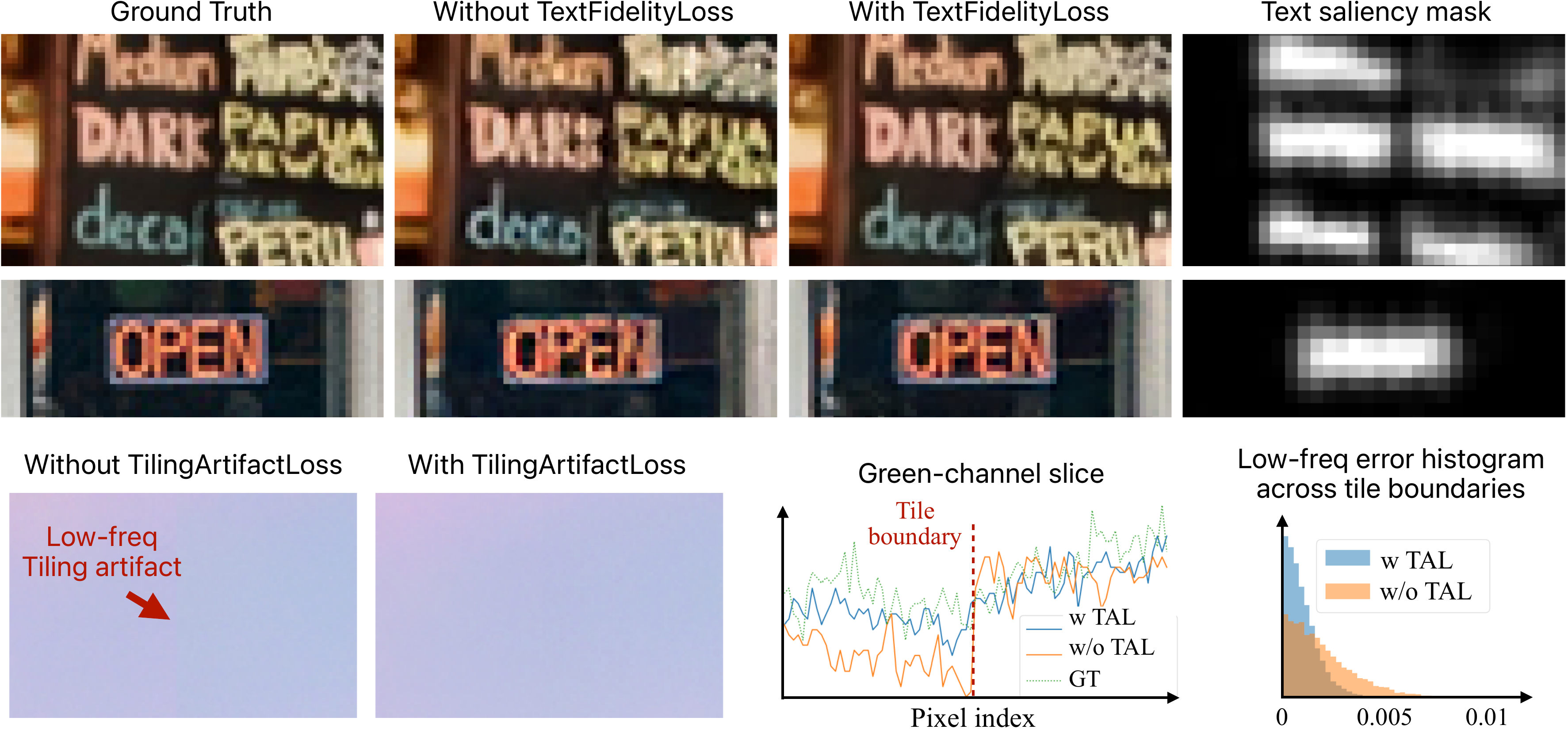}
\vspace{-0.25in}
\end{center}
   \caption{Ablations on artifact-specific mitigation strategies. All images are encoded at a BPP 0.20. \textbf{Top}: Perceptual training leads to distortions in text, while adding the TextFidelityLoss enhances its fidelity. On the right we show the text saliency masks. \textbf{Bottom}: TilingArtifactLoss (TAL) mitigates color mismatch artifacts at tile boundaries (please zoom-in to better visualize). In the middle we show a slice in the green channel across the tile boundary, where reconstruction without TAL exhibits discontinuity. On the right we show a histogram of error around tile boundaries over the CLIC 2020 Professional Validation Set, which is significantly reduced by TAL.}
\label{fig:ablation_pq}
\vspace{-0.15in}
\end{figure*}

\vspace{-0.15in}
\paragraph{Subjective study} We conduct a large-scale subjective study using Mabyduck \cite{Mabyduck}, an independent external platform for user preference studies. The study consists of pairwise blind A/B image comparison against a reference image, adopting the same standardized evaluation methodology employed by the CLIC compression challenge~\cite{naderi2023crowdsourcingapproachvideoquality, clic-challenge}. We evaluate on the CLIC 2020 Test, Kodak, and DIV2K datasets and collect a total of 74,925 pairwise comparisons from 610 unique reviewers, independently screened by Mabyduck to assure quality. Bayesian Elo scores \cite{caron2010efficientbayesianinferencegeneralized, Mabyduck-elo} are computed for each quality level of each codec based on all the pairwise comparisons, as reported in Figure~\ref{fig:perceptual_rd_curves}. Extended description of the methodology can be found in Appendix~\ref{app:subjective_study}. 

\vspace{-0.15in}
\paragraph{Speed benchmarks} We report all baseline speed numbers as quoted in their original papers or repositories, other than the iPhone runtimes. For these, we implement the exact neural architectures of the approaches, and to ensure fair apples-to-apples comparisons, we deploy them on-device with all the optimizations we applied to PICO. We benchmark all approaches on the iPhone 17 Pro Max using the tiling strategy mentioned in Sec.~\ref{sec:codec_framework} and report the neural runtimes. For PICO, we additionally report the end-to-end runtimes including all other codec components.

\vspace{-0.05in}
\subsection{Findings}
\label{sec:findings}
\vspace{-0.05in}
\paragraph{Comparisons to baselines} We show quantitative comparisons based on subjective user studies and objective metrics in Figures \ref{fig:perceptual_rd_curves},~\ref{app:fig:curves_kodak},~\ref{app:fig:curves_div2k} with a summary in Fig.~\ref{fig:spotlight_figure}. Qualitative comparisons can be found in Fig.~\ref{fig:qualitative_comparison} and Appendix~\ref{app:visual_reconstructions}. 

We observe that \ourapproach significantly outperforms all prior traditional and learned codecs across both human ratings and perceptual quality metrics, and these gains generalize across datasets. Notably, compared with today's best standardized codecs HEIC, AV1, and VVC (VTM), \ourapproach has a BD-rate of over $-60\%$ based on human ratings, suggesting a bitrate reduction of more than 2.5× for the same quality as evaluated by viewers. \ourapproach also achieves a bitrate reduction of more than 3× as compared with BPG. The subjective Elo curves in Fig.~\ref{fig:perceptual_rd_curves} also suggest that HiFiC \cite{mentzer2020high}, MRIC \cite{agustsson2023multi}, and  C3-WD \cite{balle2025good} are the three codecs which are the closest to \ourapproach with respect to compression performance. However, they are all significantly slower and less practical, while achieving 20-40\% larger file sizes for the same quality (Fig.~\ref{fig:spotlight_figure}). In general, we observe that codecs employing GANs or diffusion significantly perceptually outperform those without (\eg JPEG-AI \cite{jpegaiRepo}, MLIC++ \cite{jiang2023mlicpp}).

Qualitatively (see Fig.~\ref{fig:qualitative_comparison}), \ourapproach preserves considerably more detail than all other codecs, and produces more faithful reconstructions as compared with the original. More reconstruction examples are provided in Appendix~\ref{app:visual_reconstructions}. 

\vspace{-0.15in}
\paragraph{Network architecture ablations} 
We conduct systematic network architecture ablations, as shown in Tab.~\ref{tab:network_architecture_ablation}, to isolate the contribution of each component to the overall compression performance. The benefit of adapting quantization width to local content is evident, as its removal results in a BD-rate increase of 8.16\%. Similarly, replacing standard convolutions with our proposed ConvScale layers yields better stability and expressivity at no extra inference cost, while adding learned scaling provides additional performance gains~---~removing both results in a BD-rate increase of 9.58\%. In our one-shot context model ablations, we find that removing the component altogether causes a large performance drop of 10.28\%. Spatial AR strategies such as 2×2 grids or checkerboards deliver large improvements with minimal decoding overhead, while the minimal gains from purely channel-wise AR suggest that spatial dependencies are rather more important to capture. Conv+Haar resampling emerges as the most effective strategy for downsampling and upsampling, outperforming both pixel shuffle/unshuffle and stride-2 convolutional alternatives, while introducing no additional computational cost. Removing all ablated properties results in a BD-rate degradation of 31.69\%.

\vspace{-0.15in}
\paragraph{Artifact mitigation ablations} 
\label{sec:pq_ablation}
We ablate on the text and tiling artifact mitigations. As shown in Fig.~\ref{fig:ablation_pq} top, decoded texts are not legible in the baseline, while adding TextFidelityLoss enhances text fidelity. For a quantitative comparison, we use a test set with~$\sim100$ images with small texts. We use the same text detector~\cite{baek2019character} to label text regions, which are human-verified, and then calculate the absolute error within them (Tab.~\ref{tab:loss_ablation}). The model trained with TextFidelityLoss achieves $2\times$ lower error. In Fig.~\ref{fig:ablation_pq} bottom, we show that when TilingArtifactLoss (TAL) is missing in the training recipe, low-frequency color values visibly mismatch across tile boundaries. On the right, we show a histogram of errors across tile boundaries. The model trained with TAL has more than $2\times$ lower cross-tile error (Tab.~\ref{tab:loss_ablation}).

\vspace{-0.1in}
\section{Conclusion} 
\vspace{-0.1in}
In this work, we introduce PICO, a new image codec designed for real-life use and optimized specifically for high perceptual quality. It is the product of systematic explorations of various architectural and training recipe choices, coupled with an architecture search over millions of backbone candidates to identify models that achieve optimal tradeoffs between speed and quality. 

\newpage

\bibliographystyle{IEEEtran}
\bibliography{main} 

\clearpage
\input{appendix}
\end{document}

%% file: appendix.tex
\appendix

\section{Additional evaluations}
\label{app:additional_evaluations}
\begin{figure}[h]
    \centering
    \includegraphics[width=1\linewidth]{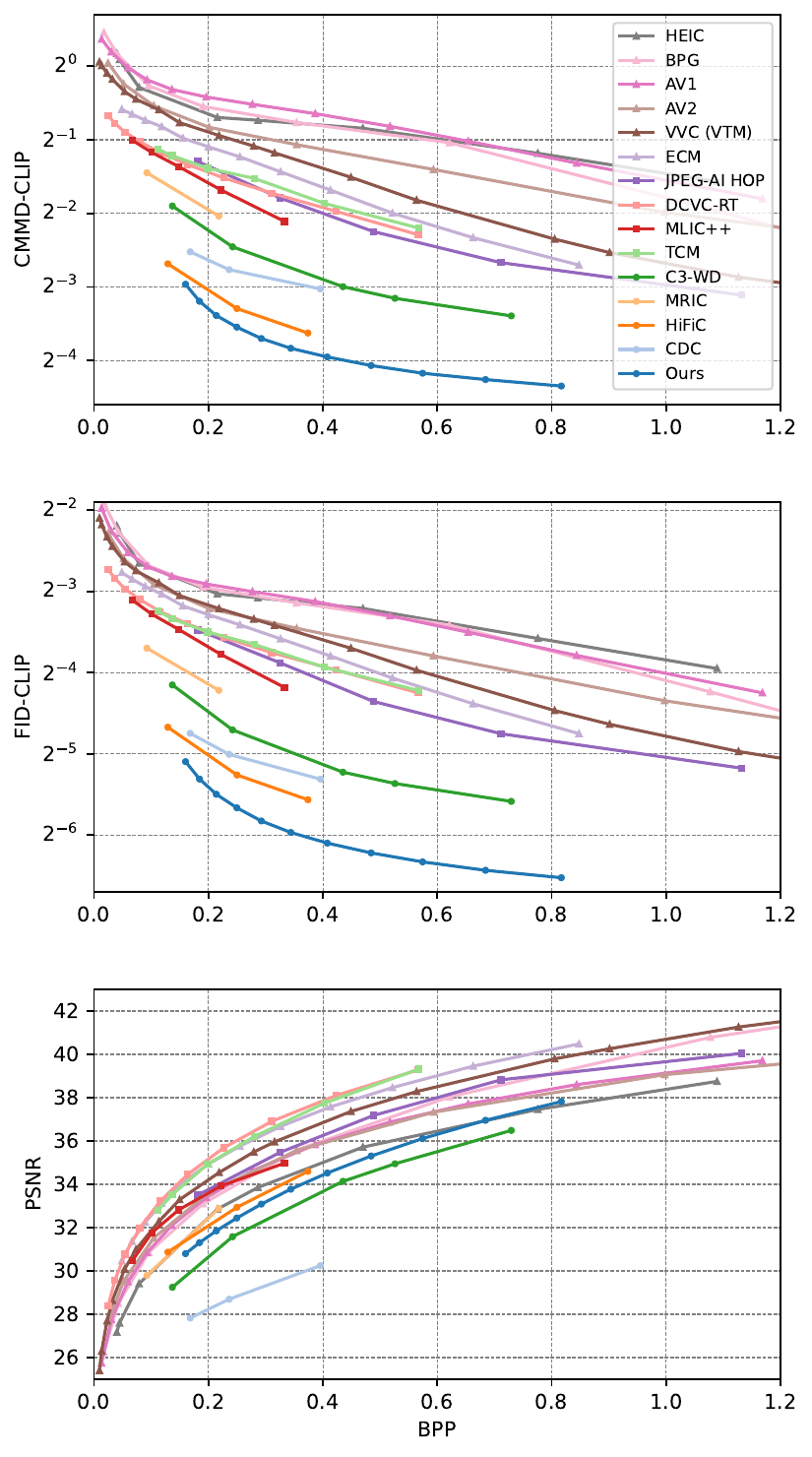}
    \caption{R-D curves for additional metrics.}
    \label{app:fig:additional_plots}
\end{figure}

Figure \ref{app:fig:additional_plots} presents curves for additional metrics. Although the perceptual codecs PICO, HiFiC, C3-WD and CDC substantially outperform the non-perceptual codecs based on human ratings and the perceptually-oriented objective metrics, they do not perform well on PSNR. Conversely, the best-performers on PSNR --- DCVC-RT, TCM, ECM, and VVC --- perform poorly on perceptual metrics, and require 2-3 times the bitrate to achieve the same perceptual quality as evaluated by viewers.

This further validates the well-known observation that PSNR poorly reflects the human visual system, and that optimizing for it comes in inherent contention with producing reconstructions that humans find to be visually faithful to the originals.

\begin{figure*}[h]
    \centering
    \includegraphics[width=1\textwidth]{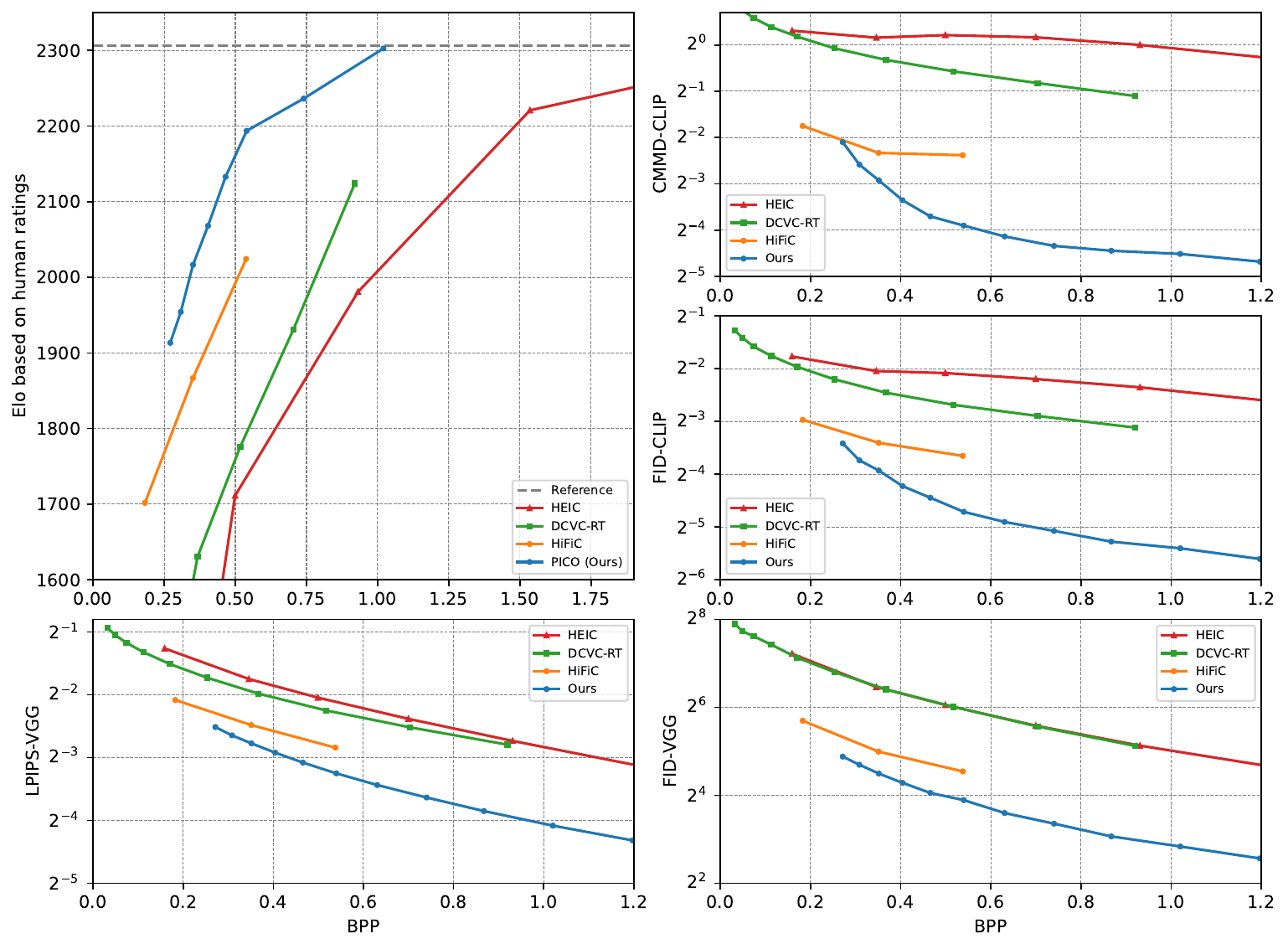}
    \caption{Subjective and objective curves for the Kodak dataset.}
    \label{app:fig:curves_kodak}
\end{figure*}

\begin{figure*}[h]
    \centering
    \includegraphics[width=1\textwidth]{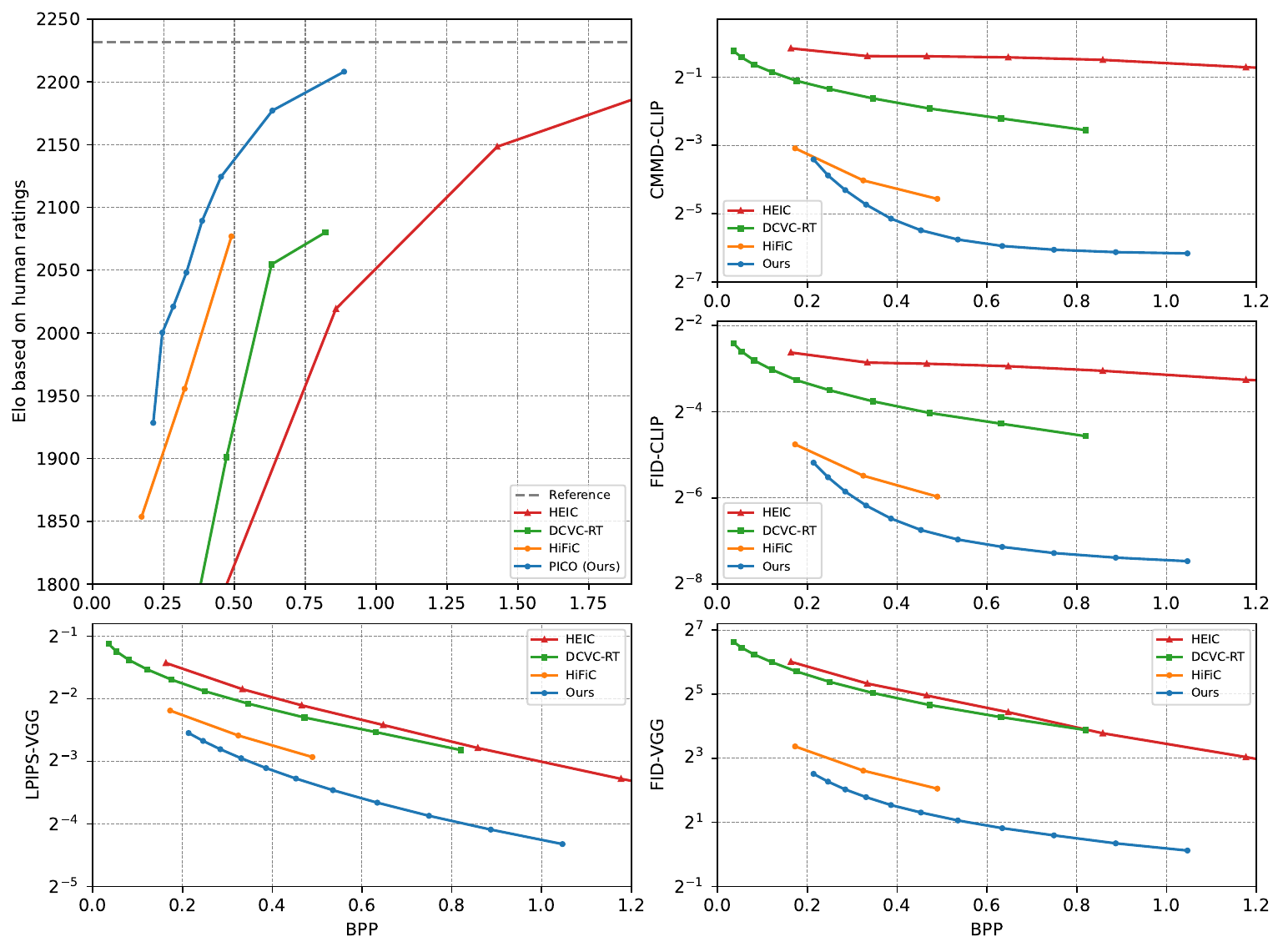}
    \caption{Subjective and objective curves for the DIV2K dataset.}
    \label{app:fig:curves_div2k}
\end{figure*}

Figures \ref{app:fig:curves_kodak} and \ref{app:fig:curves_div2k} present objective metric curves, as well as Elo curves from the subjective studies for additional evaluation datasets, Kodak and DIV2K. These showcase that PICO's subjective favorability holds across various datasets.

\section{Full model architecture}
\label{app:architecture}
\begin{figure*}[t!]
    \centering
    \includegraphics[width=1\linewidth]{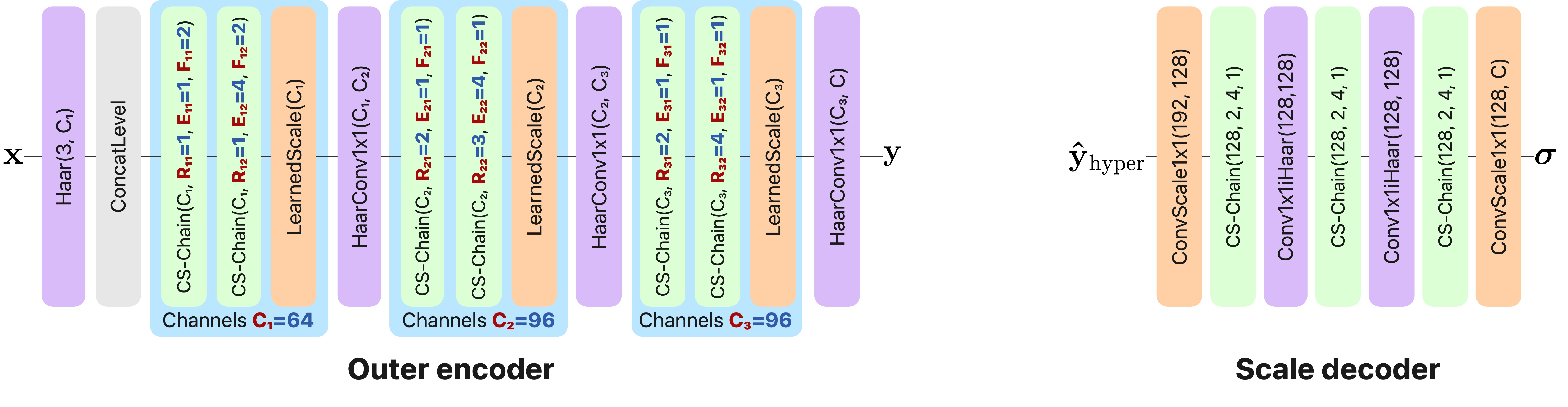}
    \caption{Architectures of the outer encoder and scale decoder. See the main paper body for details.}
    \label{fig:other_architectures}
\end{figure*}

The architectures of other parts of the model can be found in Figure \ref{fig:other_architectures}. To derive the hyperparameters of the encoder, neural architecture search was applied in a similar manner to the one of the outer decoder described in the main paper; see Section \ref{app:nas} for details.

In general, all $3\times 3$ convolutions and ConvScales in the paper are configured to have the number of channels per group be 32, unless stated otherwise.

\section{Perceptual training recipe}
\label{app:pq_training}
The training procedure is split into two phases. In the first training phase, we optimize solely for MSE distortion. The learning rate is set to 0.0008 and decayed to 30\% and 10\% of its initial value at 70\% and 90\% of training, respectively. In the second phase of the training, we introduce perceptual and GAN losses. The learning rate is decayed to 50\%/30\%/10\% of the initial rate at 30\%/60\%/80\% of training, respectively.


\section{Neural architecture search}
\label{app:nas}
Here we list the detailed search space and the chosen value for both outer encoder and outer decoder in Table \ref{HP_encoder}.
Among the models we ended up with first phase training as described in 4.3, we ranked the models with respect to their compression performance and did a thorough analysis on the impact of different hyperparameters. Take the outer decoder as example, we noticed that under the same runtime budget, putting higher channel number in the low resolution layers (i.e., scale 1) while sacrificing channel number in high resolution layers (i.e., scales 2 and 3) usually gives more benefits than other changes (such as repeat nums, 3x3 and 1x1 expansions etc). Note that although the NAS experiments were conducted on iPhone 16 Pro, we cross-validated that the conclusions generalize across different devices, including newer models, like the iPhone 17 Pro on which we reported final runtimes.

\section{Baseline codec specifications}
\label{app:baselines}

\noindent BPG \cite{bpg_source_code} encode command: 
\begin{verbatim}
bpgenc <src> \
  -q <qp> \
  -o <enc>
\end{verbatim}
The core codec underlying BPG with this distribution uses x265.

\noindent AV1 \cite{av1_source_code} encode command: 
\begin{verbatim}
aomenc <src> \
  -o <enc> \
  --cq-level=<rate> \
  --end-usage=q \
  --i420
\end{verbatim}

\noindent AV2 \cite{av2_source_code} encode command:
\begin{verbatim}
aomenc <src> \
    -o <enc> \
    --qp=<qp> \ 
    --psnr \
    --obu \
    --passes=1 \
    --end-usage=q \
    --kf-min-dist=0 \
    --kf-max-dist=0 \ 
    --use-fixed-qp-offsets=1 \
    --deltaq-mode=0 \
    --enable-tpl-model=0 \
    --cpu-used=8 \
    --enable-keyframe-filtering=0 \
    --i420
\end{verbatim}
Note that we benchmarked the AV2 reference implementation: it is the strongest baseline, but is slow and unoptimized (the reference implementations of VVC/ECM were the same or slower). 

\noindent VVC \cite{vvc_source_code} encode command:
\begin{verbatim}
EncoderAppStatic -i <src> \
  -c encoder_intra.cfg  \
  -b <enc> \
  -q <qp> \
  --ReconFile /dev/null \
  -fr 1 \
  -f 1 \
  -cf 420
\end{verbatim}

\noindent ECM \cite{ecm_source_code} encode command: 
\begin{verbatim}
EncoderAppStatic -i <src> \
  -c encoder_intra.cfg
  -b <enc> \
  -q <qp> \
  --ReconFile /dev/null \
  -fr 1 \
  -f 1 \
  --CTUSize=256
\end{verbatim}

\noindent Conversion from RGB to YUV:
\begin{verbatim}
    ffmpeg -y -loglevel quiet -i 
    "<src>" <pad_option>'-pix_fmt 
    yuv420p "<dst>"'
\end{verbatim}

\section{Quality level control}
\label{app:quality_level_control}
We start with 8 coarse levels $l_c\in 0,\ldots, 7$, which we map to one-hot vectors. We then expand the number of levels to 71, by increasing the level density 10-fold and interpolating the one-hot vector for intermediate levels between the coarse ones. Differently from \cite{rippel2021elf}, we apply the level embedding interpolation both during training and inference, rather than as just a post-training step. We condition the encoder and decoder by concatenating to their inputs the interpolated 8-dimensional one-hot tensor broadcasted spatially; we furthermore wrap the latent $\rmbhy$ with a learned level-conditional channel-wise gain and its inverse. During training, we uniformly sample quality levels, and associate a different Lagrange multiplier $\lambda_l$ for each. We further add a multiplier $\alpha_l$ reweighing each loss term as function of the level, allowing balancing the gradient during training as it accumulates across different levels. Thus, the total training loss is a combination of the distortion loss $D$ and rate loss $R$ where the latents $\rmbhy$ and reconstruction $\rmbhx$ are conditioned on the level:
\begin{eqnarray}
\label{eqn:total_loss}
\mathcal{L} = \bbE_l\left[\alpha_l D\left(\rmbx, \rmbhx_l\right) + \alpha_l\lambda_l R(\rmbhy_{\textrm{hyper}}^l, \rmbhy^l)\right]\;
\end{eqnarray}

\section{Subjective study methodology}
\label{app:subjective_study}
\begin{figure*}[t!]
    \centering
    \vspace{1in}
    \includegraphics[width=1\linewidth]{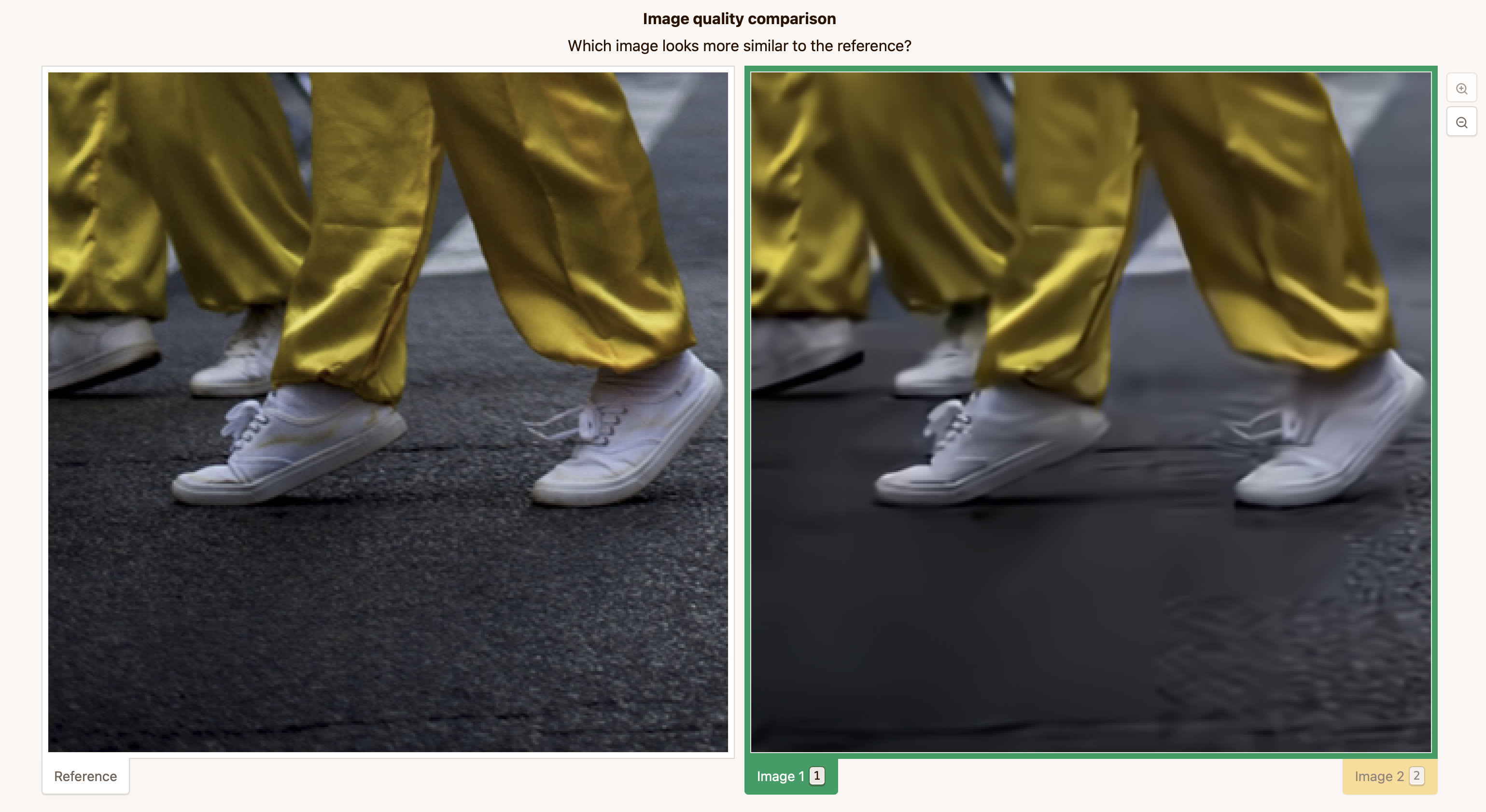}
    \caption{Screenshot of the subjective study interface as seen by the human raters}
    \label{fig:mabyduck}
\end{figure*}

The subjective study is conducted in a blind pairwise comparison format. Figure \ref{fig:mabyduck} shows the interface seen by the human raters. The interface allows for zooming, with default zoom level set to $2x$. 

Similar to the CLIC compression challenge~\cite{clic-challenge}, the study actively chooses which pair of reconstructions (corresponding to a codec evaluated at a particular rate) are compared against each other using the maximum information gain strategy \cite{Mabyduck-strategies} to maximize comparisons which provide a useful signal. Finally, Bayesian ELO scores are computed based on all the pairwise comparisons. 

To avoid noisy voting, Mabyduck performs thorough sanity checking of the reviewers setup with a pre-screening in accordance with the Ishihara color test \cite{ishihara1917}. The pre-screening checks for color blindness, contrast sensitivity and basic ability to detect compression artifacts.  A sample screening study is shown here: \url{https://xp.mabyduck.com/en/latest/pre_screen_image/job/j6ne0x2/}.  

\begin{table*}[t]
\centering
\footnotesize
\begin{subtable}[t]{0.45\linewidth}
    \centering
    \begin{tabular}{ccccc}
    \noalign{\hrule height 1pt}
    \multicolumn{3}{c|}{\makecell[c]{Hyperparameter}} & Search Space & Final value \\
    \noalign{\hrule height 1pt}
    \multirow{7}{*}{\makecell[l]{Scale 1}} 
    
    & channels
    &{$C_1$} & [32, 64] & 64 \\
    \cline{2-5}
    & \multirow{3}{*}{1st CS-Chain}
    & {$R_{11}$}  & [1, 2] & \ 1 \\
    \cline{3-5}
    & & {$E_{11}$}  & [1] & \ 1 \\
    \cline{3-5}
    & & {$F_{11}$}  & [1, 2] & \ 2 \\
    
    \cline{2-5}
    & \multirow{3}{*}{2nd CS-Chain}
    
    & {$R_{12}$}  & [1, 2] & \ 1 \\
    \cline{3-5}
    & & {$E_{12}$}  & [1, 2, 4] & \ 4 \\
    \cline{3-5}
    & & {$F_{12}$}  & [1, 2] & \ 2 \\
    
    \cline{1-5}
    \multirow{7}{*}{\makecell[l]{Scale 2}}

    & channels
    &{$C_2$} & [64, 96] &  96 \\
    \cline{2-5}
    & \multirow{3}{*}{1st CS-Chain}
    & {$R_{21}$}  & [2, 4] & \ 2 \\
    \cline{3-5}
    & & {$E_{2}$}  & [1] & \ 1 \\
    \cline{3-5}
    & & {$F_{21}$}  & [1] & \ 1 \\
    
    \cline{2-5}
    & \multirow{3}{*}{2nd CS-Chain}
    
    & {$R_{22}$}  & [1, 2, 3] & \ 3 \\
    \cline{3-5}
    & & {$E_{22}$}  & [1, 2, 4] & \ 4 \\
    \cline{3-5}
    & & {$F_{22}$}  & [1, 2] & \ 1 \\
    
    \cline{1-5}
    \multirow{7}{*}{\makecell[l]{Scale 3}} 
    & channels
    &{$C_3$} & [96, 128, 160] & 96 \\
    \cline{2-5}
    & \multirow{3}{*}{1st CS-Chain}
    & {$R_{31}$}  & [2, 4, 6] & \ 2 \\
    \cline{3-5}
    & & {$E_{31}$}  & [1] & \ 1 \\
    \cline{3-5}
    & & {$F_{31}$}  & [1] & \ 1 \\
    
    \cline{2-5}
    & \multirow{3}{*}{2nd CS-Chain}
    
    & {$R_{32}$}  & [2, 4] & \ 4 \\
    \cline{3-5}
    & & {$E_{32}$}  & [1, 2, 4] & \ 1\\
    \cline{3-5}
    & & {$F_{32}$}  & [1, 2] & \ 1 \\
    
    \noalign{\hrule height 1pt}
    \end{tabular}
    \caption{Outer Encoder}
\end{subtable}
\hfill
\begin{subtable}[t]{0.45\linewidth}
    \centering
    \begin{tabular}{ccccc}
    \noalign{\hrule height 1pt}
    \multicolumn{3}{c|}{\makecell[c]{Hyperparameter}} & Search Space & Final value \\
    \noalign{\hrule height 1pt}
    \multirow{7}{*}{\makecell[l]{Scale 1}} 
    
    & channels
    &{$C_1$} & [96, 128, 160] & 160 \\
    \cline{2-5}
    & \multirow{3}{*}{1st CS-Chain}
    & {$R_{11}$}  & [2, 3, 4] & \ 3 \\
    \cline{3-5}
    & & {$E_{11}$}  & [1] & \ 1 \\
    \cline{3-5}
    & & {$F_{11}$}  & [1] & \ 1 \\
    
    \cline{2-5}
    & \multirow{3}{*}{2nd CS-Chain}
    
    & {$R_{12}$}  & [2, 3] & \ 2 \\
    \cline{3-5}
    & & {$E_{12}$}  & [1, 2, 4] & \ 1 \\
    \cline{3-5}
    & & {$F_{12}$}  & [1, 2] & \ 2 \\
    
    \cline{1-5}
    \multirow{7}{*}{\makecell[l]{Scale 2}}

    & channels
    &{$C_2$} & [64, 96] & 64 \\
    \cline{2-5}
    & \multirow{3}{*}{1st CS-Chain}
    & {$R_{21}$}  & [1, 2, 3] & \ 2 \\
    \cline{3-5}
    & & {$E_{2}$}  & [1] & \ 1 \\
    \cline{3-5}
    & & {$F_{21}$}  & [1] & \ 1 \\
    
    \cline{2-5}
    & \multirow{3}{*}{2nd CS-Chain}
    
    & {$R_{22}$}  & [1, 2] & \ 1 \\
    \cline{3-5}
    & & {$E_{22}$}  & [1, 2, 4] & \ 4 \\
    \cline{3-5}
    & & {$F_{22}$}  & [1, 2] & \ 2 \\
    
    \cline{1-5}
    \multirow{7}{*}{\makecell[l]{Scale 3}} 
    
    & channels
    &{$C_3$} & [32, 64] & 32 \\
    \cline{2-5}
    & \multirow{3}{*}{1st CS-Chain}
    & {$R_{31}$}  & [1, 2] & \ 2 \\
    \cline{3-5}
    & & {$E_{31}$}  & [1, 2] & \ 2 \\
    \cline{3-5}
    & & {$F_{31}$}  & [1, 2] & \ 2 \\
    
    \cline{2-5}
    & \multirow{3}{*}{2nd CS-Chain}
    
    & {$R_{32}$}  & [1, 2] & \ 1 \\
    \cline{3-5}
    & & {$E_{32}$}  & [1, 2, 3] & \ 3 \\
    \cline{3-5}
    & & {$F_{32}$}  & [1, 2] & \ 2 \\
    
    \noalign{\hrule height 1pt}
    \end{tabular}
    \caption{Outer Decoder}
\end{subtable}
\caption{Neural architecture search summary for outer encoder and decoder.}
\label{HP_encoder}
\end{table*}

\section{Conv + Haar resampling implementation details}
\label{app:haar}
We use Haar wavelets for all resampling operations in the codec~---~while adding zero additional computation, via a reparametrization trick. In our model, the resampling operation is always coupled with a change of the number of channels. For instance, the encoder might need to downsample by 2× from one spatial scale with $C_1$ channels, to another with $C_2$ channels. This could be achieved with applying a Haar transform, followed by a $1\times 1$ convolution mapping from $C_1\rightarrow C_2$. Observing that Haar can be expressed as a simple $4\times 4$ matrix multiplication to the 4 elements of $2\times 2$ spatial blocks, we combine the Haar and the 1x1 convolution into a single 1x1 convolution with a modified weight into which Haar is collapsed, preceded by a factor-2 space-to-depth. The decoder-side conv+iHaar upsampling operation is treated in an analogous way.

\section{Limitations}
PICO is optimized for perceptual quality specifically for \textit{natural} contents. On extremely simple synthetic contents (\eg, cartoon), PICO uses a higher bitrate compared to conventional codecs to achieve similar quality. This is because the image perfectly fits conventional codecs' autoregressive modeling. 


\section{Additional reconstructions}
\label{app:visual_reconstructions}
Reconstructions of PICO on examples from the CLIC 2020 test set can be found at \url{https://ml-site.cdn-apple.com/datasets/lic/pico.zip}. Photos of people with visible faces had to be removed due to licensing limitations.

Additional visual comparisons of PICO against HiFiC, VVC (VTM) and the original uncompressed image can be found at the end of the supplementary materials.

Multiple issues can be seen in HiFiC relative to PICO:
\begin{itemize}
    \item Over-synthesis: it hallucinates details, at the cost of fidelity to the original image.
    \item Synthesis of incorrect statistics relative to the original: it introduces patterns to smooth surfaces, and over-sharpens edges and textures.
    \item HiFiC is often unable to keep small text legible.
    \item HiFiC exhibits noticeable structured repetitive patterns, where the underlying texture is more random.
\end{itemize}

\begin{figure*}[h]
    \centering
    \includegraphics[width=1\linewidth]{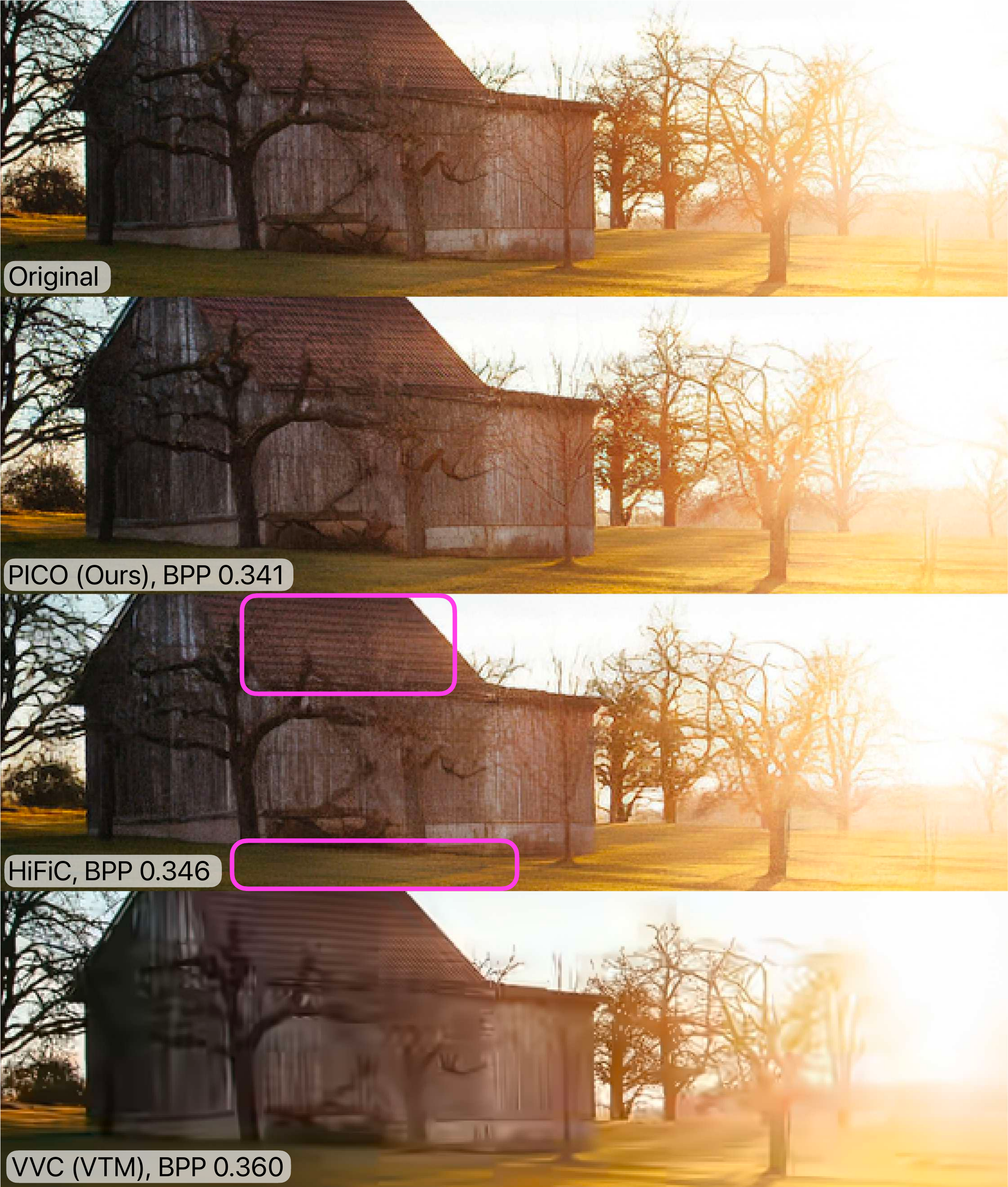}
\end{figure*}

\begin{figure*}[h]
    \centering
    \includegraphics[width=1\linewidth]{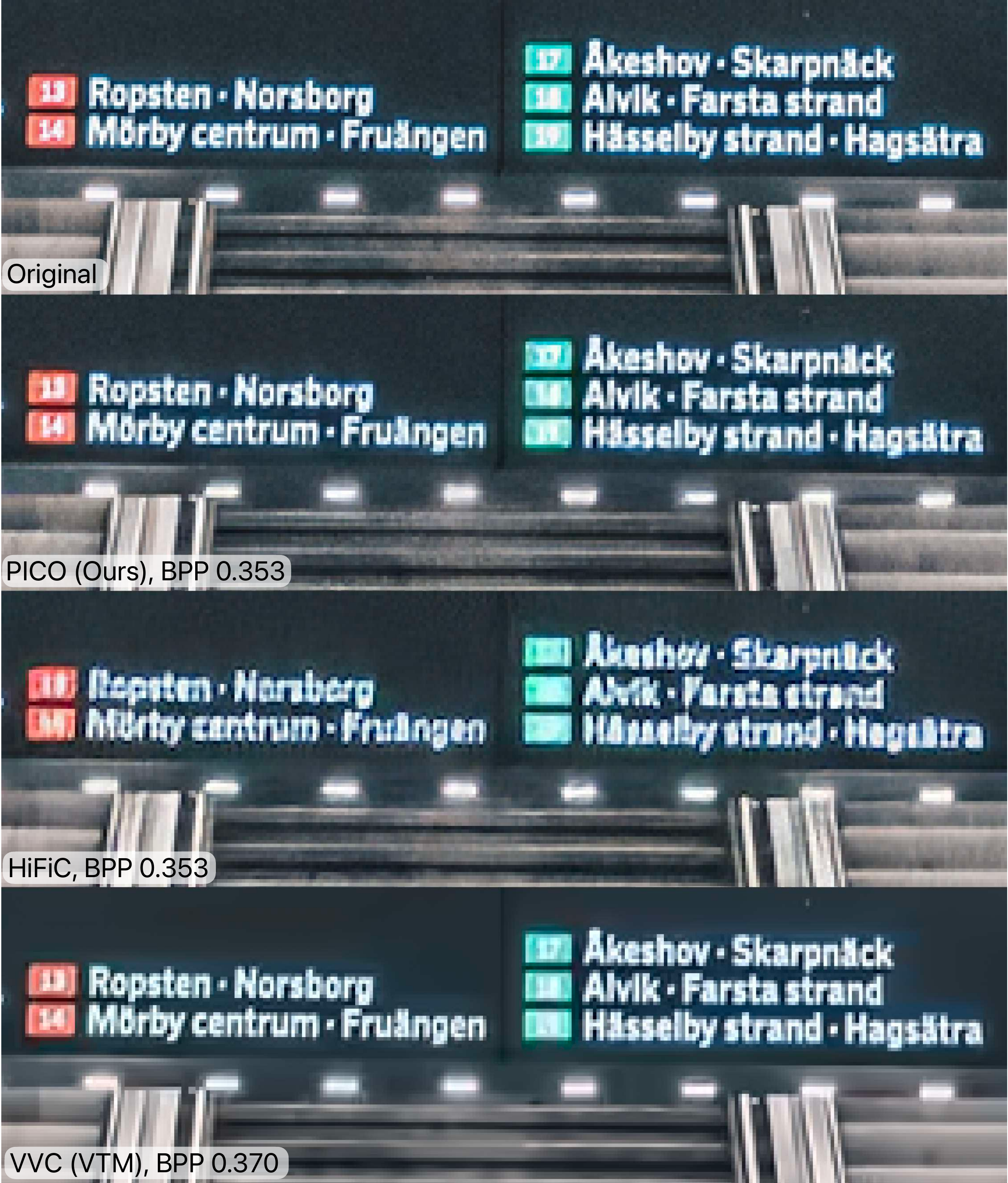}
\end{figure*}

\begin{figure*}[h]
    \centering
    \includegraphics[width=1\linewidth]{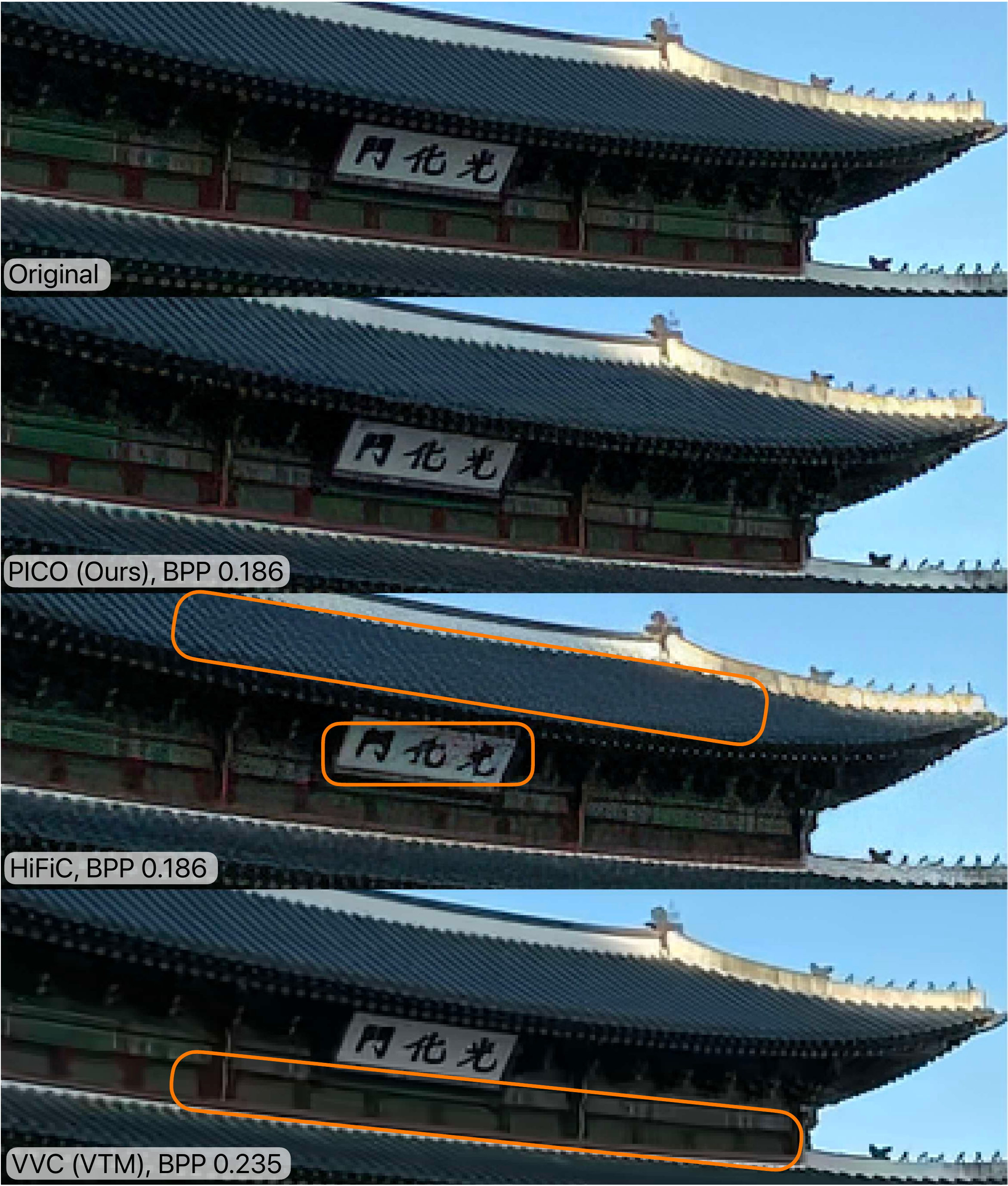}
\end{figure*}

\begin{figure*}[h]
    \centering
    \includegraphics[width=1\linewidth]{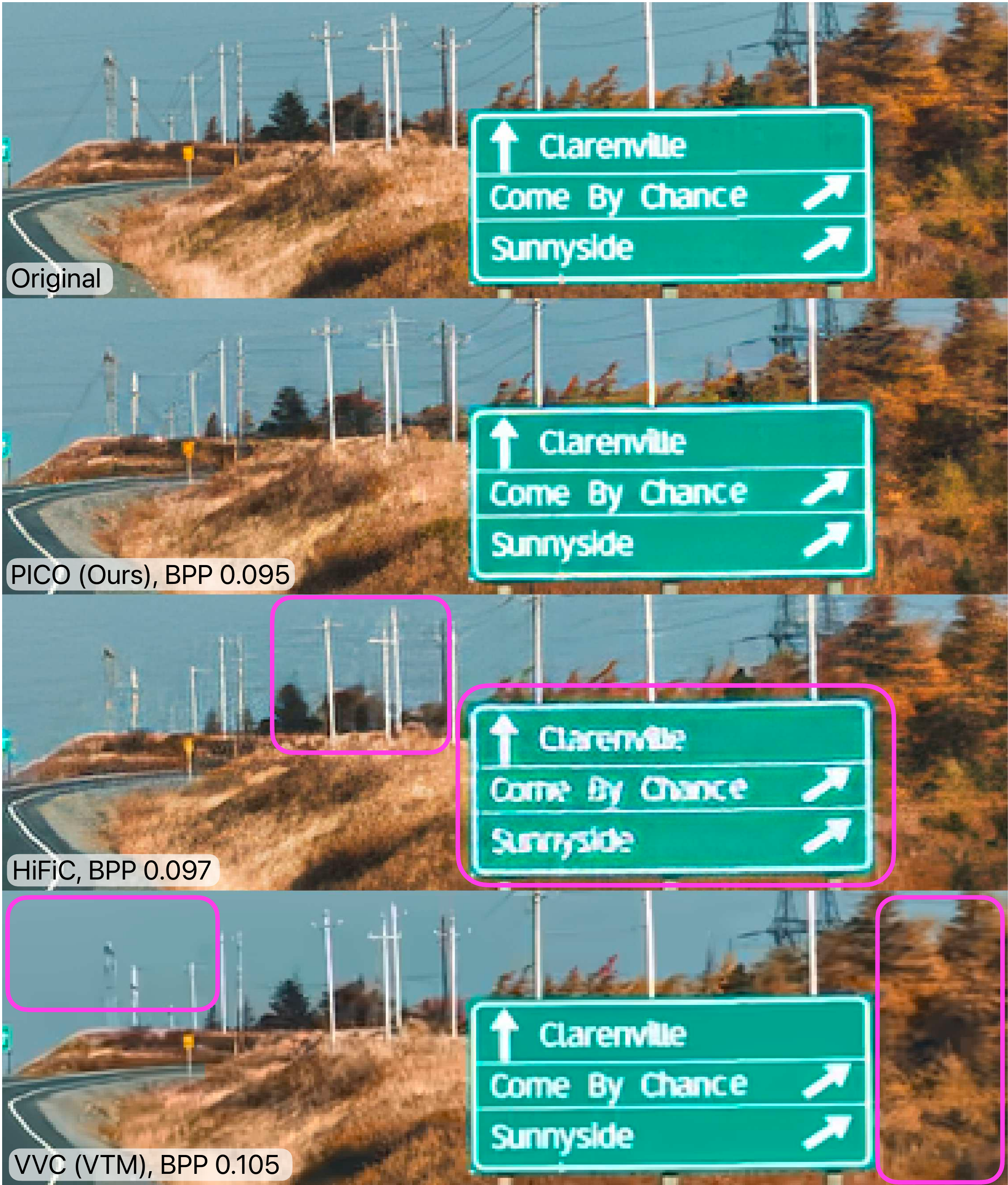}
\end{figure*}

\begin{figure*}[h]
    \centering
    \includegraphics[width=1\linewidth]{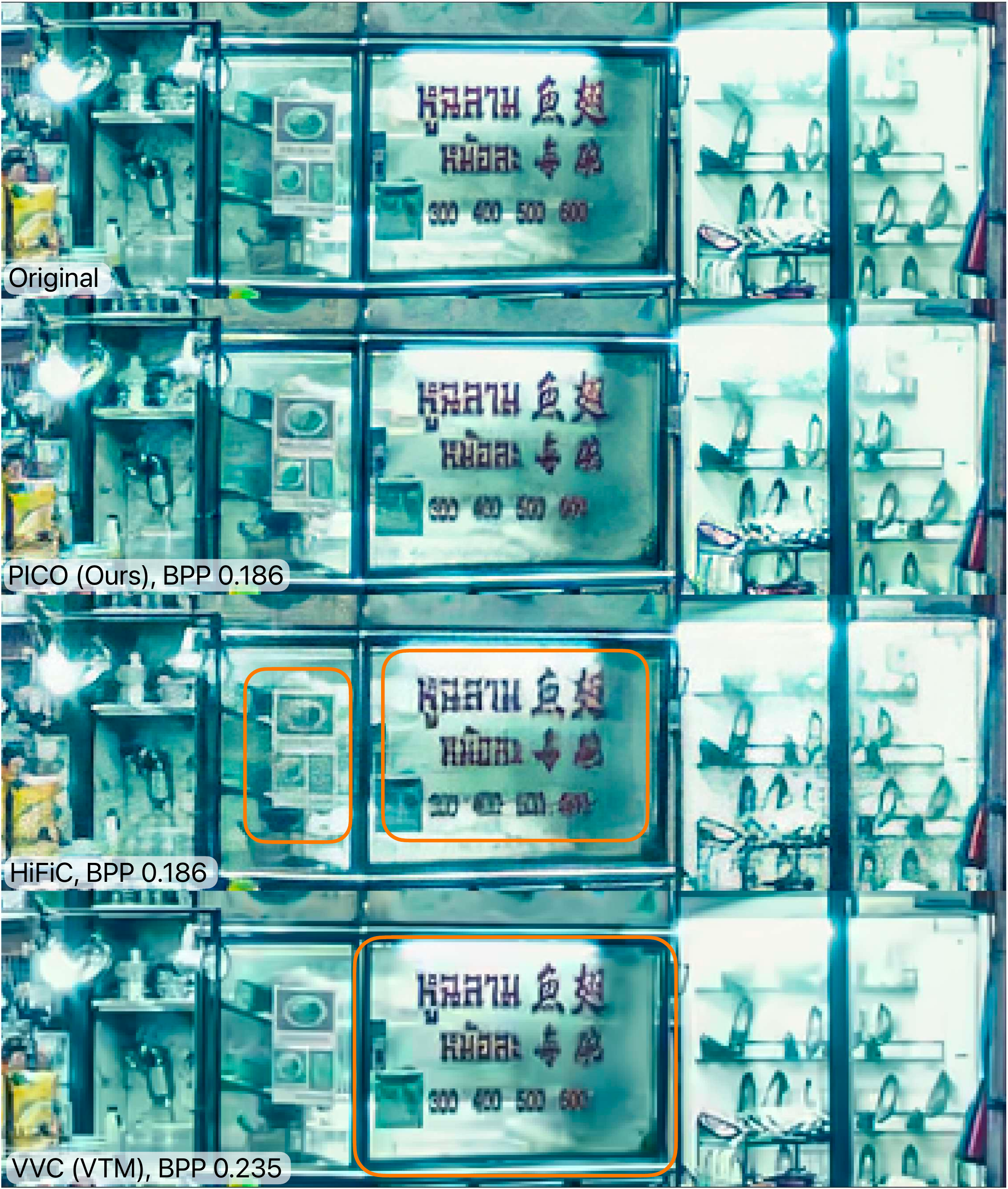}
\end{figure*}

\begin{figure*}[h]
    \centering
    \includegraphics[width=1\linewidth]{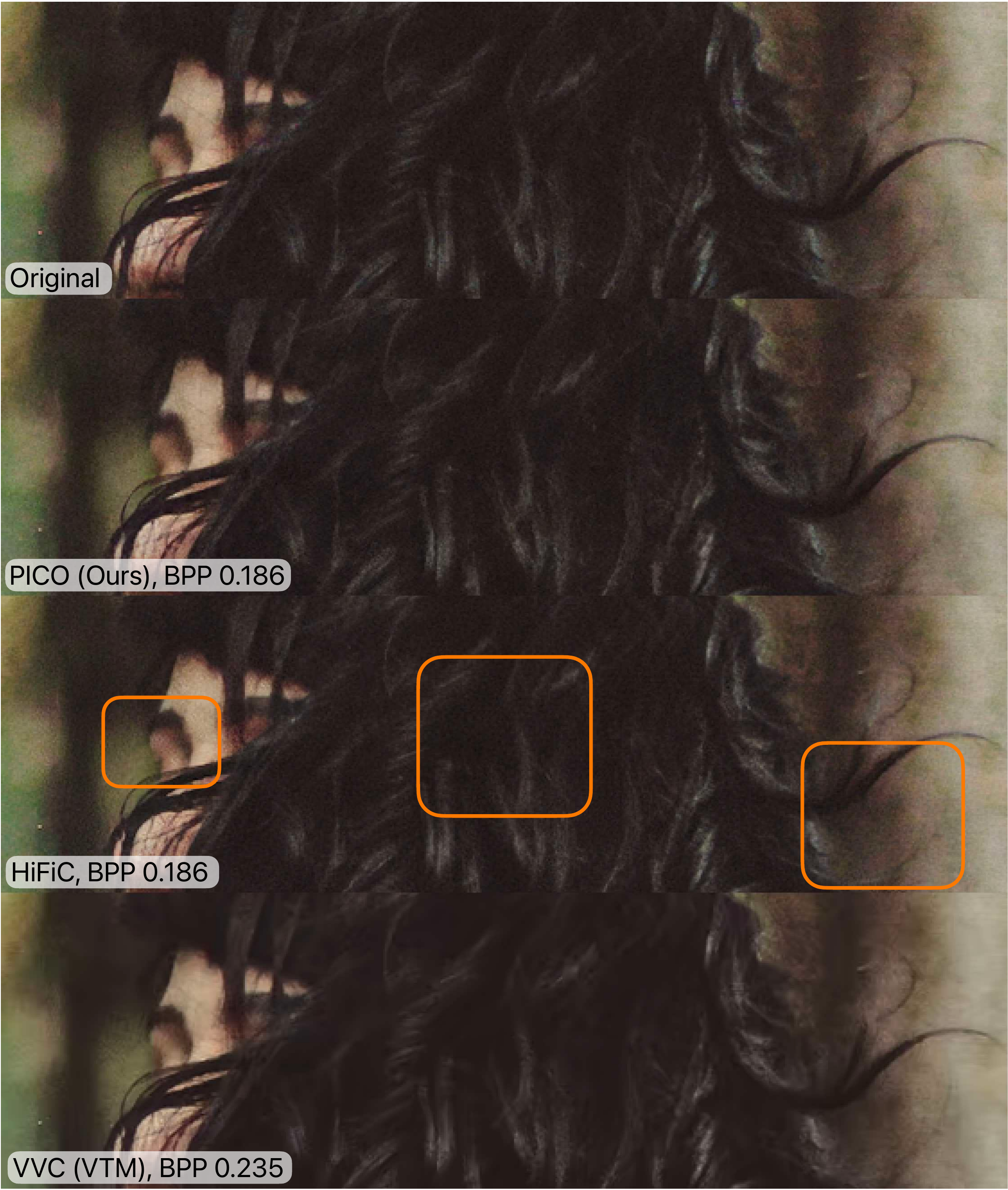}
\end{figure*}

\begin{figure*}[h]
    \centering
    \includegraphics[width=1\linewidth]{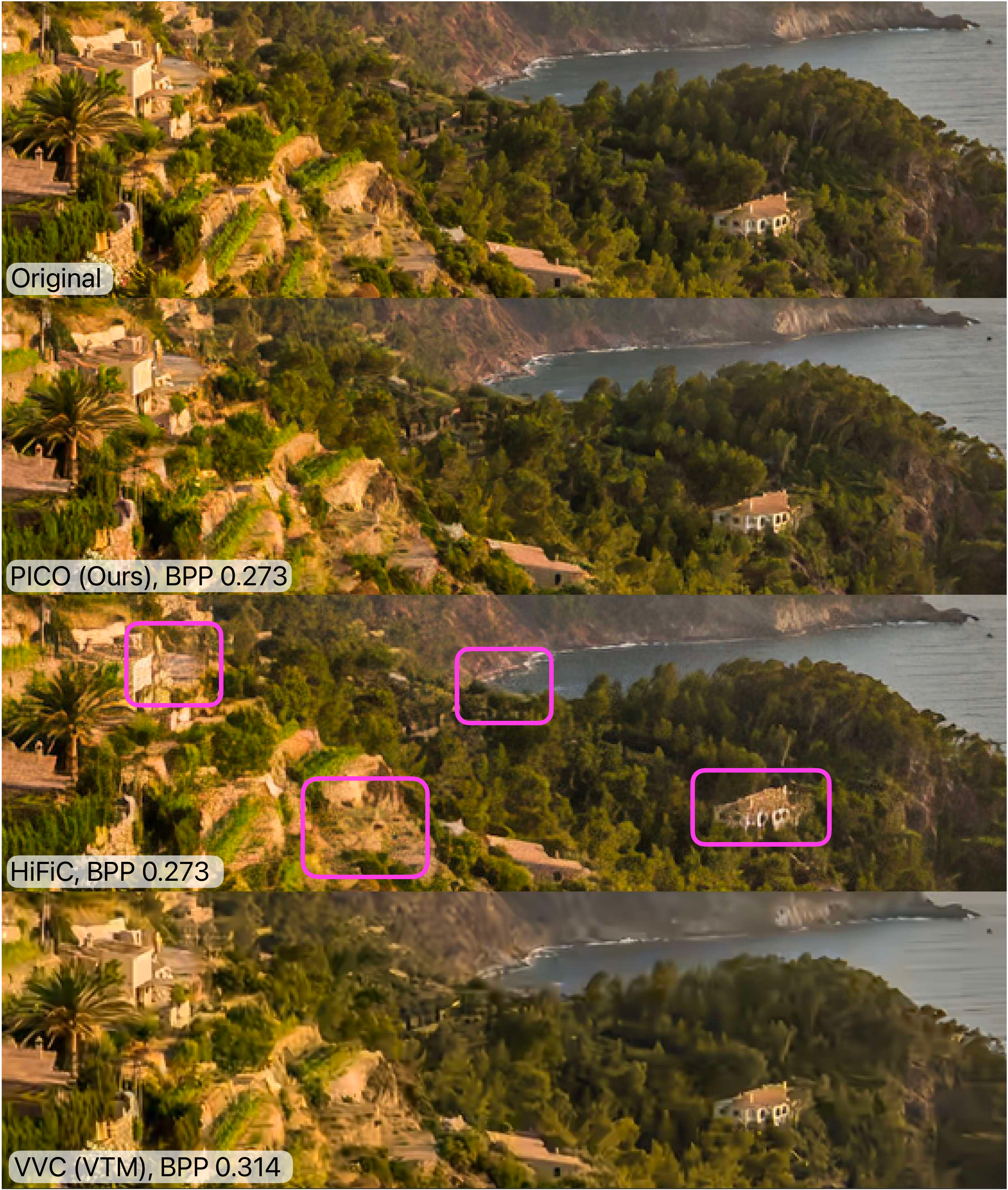}
\end{figure*}
